# Bits or permutations? Spending a fixed inference budget on weight precision or on option-order averaging in small local language models

*Simulated low-bit quantization, option-order sensitivity and a compute-matched exchange rate (Qwen3-0.6B and Qwen3-1.7B, four multiple-choice benchmarks)*

Mohammed Nayeem Khan
mohammed.nayeem.khan@intel.com
mnknayeem@gmail.com

Logesh L
logesh.l@intel.com

Aliyas Ahmad Misri
aliyas.ahmad.misri@intel.com

Intel Corporation

**Abstract**

Local language models can save memory by storing weights in fewer bits, and they can reduce sensitivity to the order of answer options by averaging a few reorderings. We ask whether the first makes the second worse, and which is the cheaper way to spend a fixed budget. On Qwen3-0.6B and Qwen3-1.7B with simulated round-to-nearest weight quantization (8, 6, 4 and 3 bits, fp16 reference), read from the logits of the letters A to D on ARC-Easy, ARC-Challenge, MMLU and HellaSwag (300 items each), 6 bits lose no accuracy against fp16 (they score +1.8 (+1.0 to +2.5) and +0.9 (+0.3 to +1.5) points higher, which we do not explain), and 8 bits change it by -0.7 (-1.1 to -0.3) and -0.4 (-0.7 to -0.2) points. At 4 bits the answer-slot skew rises from 0.15 to 0.53 (0.6B) and from 0.07 to 0.12 (1.7B); Gaussian noise at the same weight SNR gives 0.15 and 0.37 (two draws) and 0.19 (one draw), so the control is inconclusive on whether rounding adds slot skew beyond equal-power noise. Flip rates rise but stay below an accuracy-matched baseline of independent uniform errors. Averaging four cyclic orders recovers +12.6 points at 4 bits for 0.6B (+5.3 at fp16). We price bits and passes in one machine-independent unit, weight bytes read times passes. In cost-matched pairs, extra orders beat extra bits only when the weights have 6 or more bits, and the decision table reduces to: use 6-bit weights and spend the rest of the budget on up to four orders, one pass each. Everything is simulated on two small models with the simplest quantizer; a title-level search found no earlier study of quantization together with option order.

**Key findings (every number recomputed from the cached per-item logits)**

1. 6-bit round-to-nearest loses no accuracy against fp16 on either model (6 bits scores higher by +1.8 (+1.0 to +2.5) and +0.9 (+0.3 to +1.5) points, intervals excluding zero), and 8 bits change it by -0.7 (-1.1 to -0.3) and -0.4 (-0.7 to -0.2) points. We did not find the reason for the 6-bit gain and do not claim that 6 bits improve a model. 6 bits read 0.39 of the fp16 weight bytes.
2. At 4 bits the answer-slot skew is 0.53 (0.6B) and 0.12 (1.7B) against 0.15 and 0.07 at fp16. Matched-SNR Gaussian noise gives 0.15 and 0.37 (two draws) and 0.19 (one draw), so the control is inconclusive on whether rounding itself adds slot skew.
3. Flip rates rise with lost bits but the 4-bit excess over an accuracy-matched baseline is -5.1 (-6.2 to -4.0) points (0.6B) and -5.3 (-7.5 to -2.9) (1.7B). Under this baseline, which assumes independent uniform errors, accuracy loss accounts for the flips, while the slot skew lies above it.
4. Averaging four cyclic orders adds +5.3 points at fp16 and +12.6 at 4 bits for 0.6B (+3.8 and +4.9 for 1.7B) and cuts the ECE from 0.33 to 0.09 (0.6B, 4 bits).
5. In the cost-matched pairs of Table 6 (a many-order cell at 0.5 to 1.13 times the cost of a one-order cell), extra orders win on accuracy in 2 of 10 comparisons for 0.6B and 2 of 10 for 1.7B, always with a fewer-bit cell of 8 or 6 bits. More bits win in 7 and 7 comparisons, with a fewer-bit cell of 4 or 3 bits. The accuracy-cost frontier of both models consists of 6-bit cells with K = 1 to 4 plus the cheapest 3- and 4-bit single passes (Section 10).
6. Decision rule from the cross-fitted table: with a budget of at least 0.391 fp16 passes use 6-bit weights and K = floor(B / 0.391) orders up to 4; below that use 4 bits and one order (weights only, no fp16 embedding cost).
7. Fragile questions are not stable across precisions: the per-question flip share of fp16 and 4-bit correlates at 0.20 (0.6B) and 0.64 (1.7B). Two equal-SNR noise draws of the 0.6B model correlate at 0.15, so the instability is specific to each perturbed copy of the model.

## 1 Introduction and short answer

People who run a language model on their own machine have two cheap ways to trade quality against effort. They can store the weights in fewer bits, and they can run the model more than once. In multiple-choice answering the two interact, and we have not seen that measured. Reordering the answer options changes the answer of many small models, and quantization changes the model that is being reordered. If low-bit weights make a model more sensitive to option order, the standard remedy (answer under several orders and average) pays twice. A user with a fixed budget then has to decide how many bits to give up for how many extra passes.

We study two questions on Qwen3-0.6B and Qwen3-1.7B with simulated weight quantization.

**Question 1.** Does low-bit round-to-nearest quantization increase order sensitivity (answer flips under option permutations) and the slot prior (a preference for one answer letter) beyond what the loss of accuracy explains? The bit widths are 8, 6, 4 and 3, with one fp16 scale per 64 weights and fp16 as reference.

**Question 2.** At equal cost, does it help to quantize harder and average K cyclic option permutations, compared with keeping more bits and answering once? Cost is machine independent: weight bytes read, summed over forward passes, in units of one fp16 pass.

Both questions are read from the logits of the letters A to D after a fixed prompt on ARC-Easy, ARC-Challenge, MMLU and HellaSwag (300 items per task and model, 1,200 per model). Intervals are item-level bootstrap 95% intervals, and all comparisons are paired on the same items. A control perturbs the weights with Gaussian noise at the same per-group signal-to-noise ratio as the quantizer, which separates the effect of rounding from the effect of any weight error.

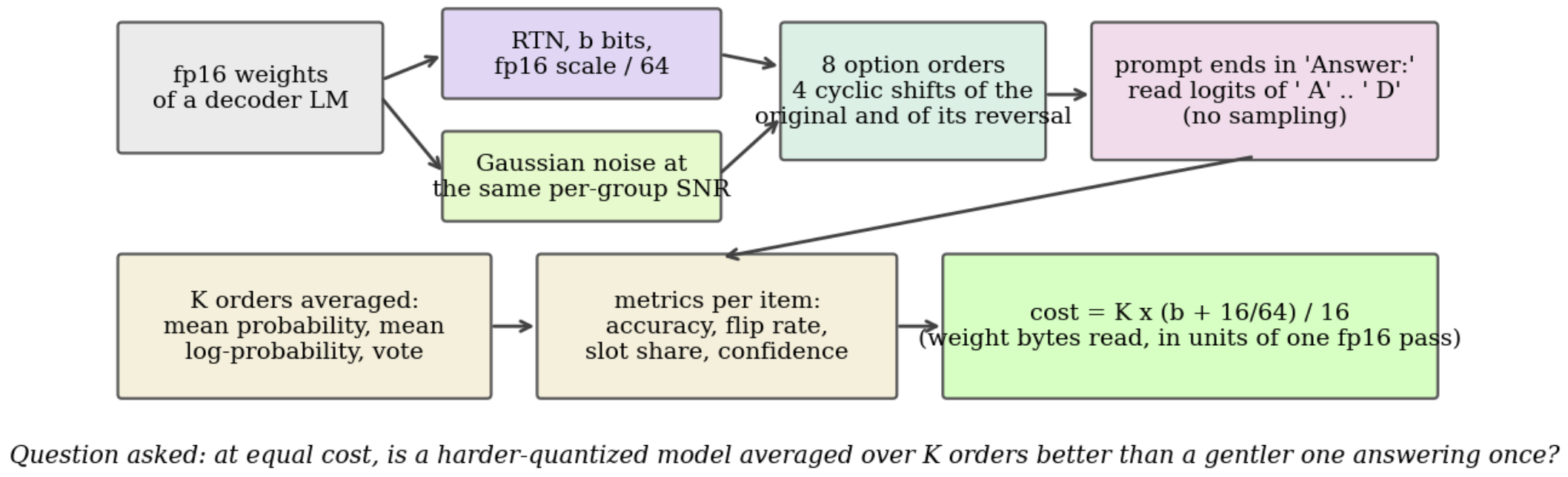

**Figure 1.** The study in one picture. Weights are perturbed in memory (round-to-nearest at b bits, or Gaussian noise with the same per-group error power). Each question is shown under eight option orders and the model is read through the logits of the four letters after a fixed prompt. Averaging K orders costs K forward passes, and every pass costs in proportion to the bits of its weights.

### 1.1 Short answers

On the slot prior (Q1), every model prefers some answer slot even at fp16 (slot skew 0.15 for 0.6B, 0.07 for 1.7B), and 6-bit and 8-bit rounding move it by at most 0.05. At 4 bits the skew is 0.53 (0.6B) and 0.12 (1.7B), so the rise is large where accuracy collapses and small where it holds. Gaussian noise at the same per-group SNR gives 0.15 and 0.37 (two draws) for 0.6B and 0.19 (one draw) for 1.7B. A perturbation of that size produces a slot prior with or without rounding, and the noise result varies between draws (Section 6).

Flip rates (Q1) rise with bit loss, but at 4 bits they stay below what an accuracy-matched baseline predicts (excess -5.1 (-6.2 to -4.0) points for 0.6B and -5.3 (-7.5 to -2.9) for 1.7B). The loss of accuracy therefore accounts for the flips under this baseline. The baseline assumes independent uniform errors, so it is a weak test, and the result is only that no excess flips were detected. The slot skew lies above the same baseline (0.53 against 0.06 for 0.6B, 0.12 against 0.06 for 1.7B).

Averaging four cyclic orders adds +5.3 points of accuracy at fp16 and +12.6 at 4 bits for 0.6B (+3.8 and +4.9 for 1.7B), so a quantized model leaves more for permutations to collect.

In the cost-matched pairs of Table 6 (a many-order cell at 0.5 to 1.13 times the cost of a one-order cell), extra orders win on accuracy in 2 of 10 comparisons for 0.6B and 2 of 10 for 1.7B, always with a fewer-bit cell of 8 or 6 bits. More bits win in 7 and 7 comparisons, with a fewer-bit cell of 4 or 3 bits. The accuracy-cost frontier of both models consists of 6-bit cells with K = 1 to 4 plus the cheapest 3- and 4-bit single passes (Section 10).

The quantizer is the simplest one and better quantizers lose less at 3 and 4 bits, so we make no comparison against any named quantizer. Everything is stated for these two models, this prompt format and these four benchmarks. As far as a title-level search of Hugging Face Papers showed, no earlier study combines quantization with option order (Section 13).

## 2 Setting: models, data, perturbations, orders and metrics

### 2.1 Models and data

We use Qwen3-0.6B and Qwen3-1.7B (Apache-2.0; revisions c1899de and 70d244c). Qwen3-4B was not run because its fp16 weights (about 8 GB) exceed the 6 GB we kept free on a shared GPU. The benchmarks are ARC-Easy and ARC-Challenge (test, four-option items), MMLU (test; a draw proportional to subject size, shuffled) and HellaSwag (validation; the four endings of each item). For every task we fix a shuffled order of items once (seed 1234) and use the first 300. Per task and model that is 300 items, and 1,200 per model when the four tasks are pooled; pooled numbers weight each task equally.

The prompt is plain text with no chat template: a one-line instruction, "Question:", the question, four lines "A. ...", "B. ...", "C. ...", "D. ...", and "Answer:". We read the logits of the four tokens " A", " B", " C", " D" at the last position and take the arg max (no sampling, no thinking phase). For HellaSwag the question is "Choose the most plausible continuation of the text:" followed by the context. Appendix D shows an example.

## 2.2 Simulated quantization and the noise control

For every linear layer in the decoder blocks (q, k, v, o, gate, up and down projections), the weights along the input dimension are split into groups of 64. In each group the scale is s = max|w| / (2^(b-1) - 1), stored in fp16; each weight becomes q = clip(round(w / s), -2^(b-1), 2^(b-1) - 1) and the model computes with s q in fp16. Embeddings, the tied output matrix and the norms stay in fp16. This is round-to-nearest (RTN) with absmax scales, simulated in memory: nothing is packed and no low-bit kernel runs.

For the noise control, we measure the root-mean-square rounding error of the RTN weights in each group and replace the rounding error with independent Gaussian noise of that RMS. Signal-to-noise ratio is therefore matched group by group, and pooled over the model (Table 1). Noise is independent of the weight values, whereas rounding sets 16% of the 4-bit weights to exactly zero and makes its error a deterministic function of the weights. At 4 and 3 bits the total weight energy after rounding is within 0.1% and 0.4% of the energy after noise (4-bit: 1.0131 against 1.0123 times the original for 0.6B), so rounding does not simply shrink the weights. Noise is drawn once per setting, and a second independent draw was run at 4 and 3 bits for the 0.6B model (all four tasks) and at 4 bits for the 1.7B model (ARC-Easy only).

**Table 1.** Configurations. Effective bits per weight include one fp16 scale per group of 64 (16/64 = 0.25 bit). Cost is the weight bytes read in one forward pass relative to fp16, with every weight counted at the stated width (Section 14 gives the cost when the fp16 embedding matrix is counted). SNR is the pooled signal-to-noise ratio of the perturbed decoder weights in dB (undefined for the unperturbed fp16 reference); the last column is the share of 0.6B weights that round to zero.

| Configuration | Bits | Effective bits | Cost per pass | SNR 0.6B (dB) | SNR 1.7B (dB) | Rounded to zero |
|---|---|---|---|---|---|---|
| fp16 (reference) | 16 | 16.00 | 1.000 | - | - | - |
| 8-bit RTN | 8 | 8.25 | 0.516 | 44.3 | 44.2 | 0.9% |
| 6-bit RTN | 6 | 6.25 | 0.391 | 32.0 | 31.9 | 3.6% |
| 4-bit RTN | 4 | 4.25 | 0.266 | 19.1 | 19.0 | 15.8% |
| 3-bit RTN | 3 | 3.25 | 0.203 | 11.8 | 11.7 | 35.3% |
| noise at 4-bit SNR | 4 | (as RTN) | 0.266 | 19.1 | 19.0 | - |
| noise at 3-bit SNR | 3 | (as RTN) | 0.203 | 11.8 | 11.7 | - |

## 2.3 Option orders, K-order averages and metrics

Each question is shown under eight orders: the four cyclic shifts of the original order (ABCD, BCDA, CDAB, DABC in terms of the original options) and the four cyclic shifts of the reversed order. Within each group of four, every option sits in every slot exactly once, so the gold answer is uniform over slots by construction and a slot preference cannot be mistaken for a quirk of the answer key. We call the two groups the forward and the reversed orbit.

*Accuracy* is the share of correct arg maxes, averaged over the orders (one-pass accuracy) or computed from a K-order aggregate. *Flip rate* is the share of pairs of orders for which the chosen option differs. For K orders we aggregate the letter probabilities back to original options (mean probability, mean log-probability, or majority vote with mean-probability tie break) and pick the arg max. We then compare the decision made from K orders of the forward orbit with the decision made from K orders of the reversed orbit. With K = 1 this is a single-order flip rate across the two orbits. At K = 4 it is the order sensitivity that remains after averaging over all cyclic shifts: the average over a full orbit is invariant to cyclic reordering, so only the difference between orbits is left. We average over all C(4, K) choices of orders in each orbit.

*Slot skew* is the total variation distance between the distribution of the chosen slot (A to D) and the uniform one. The *same-slot share* is the share of pairs of orders within an orbit for which the chosen slot is the same. The orders place each option in a different slot, so a pair with the same slot has a different option. Following Ok et al. (arXiv:2610.20993), the flip rate and the slot skew are the two factors of selection bias, and we report both. *Cost* is K (b + 16/64) / 16 for b-bit weights, and K for fp16; a pass of 8-bit weights costs 0.516 and one pass of 3-bit weights costs 0.203. Appendix A gives the formulas.

### 2.4 Statistics

All intervals are percentile intervals over 1,000 bootstrap resamples of items (pooled over the four tasks unless a task is named). The same resamples are used for every quantity, so a difference between two configurations is a paired bootstrap difference. We report 95% intervals and, for differences, a two-sided bootstrap p value (twice the smaller tail share). With four tasks, two models and about ten comparisons per section we treat p values as a guide and rely on intervals and on agreement between models and tasks.

## 3 A worked example

Take ARC-Easy item 10 of the fixed draw: *Which items would be most useful at home during a power failure?*

- A. flashlights and extra batteries (gold)
- B. hats and sunscreen
- C. raincoats and umbrellas
- D. insect spray and jackets

The fp16 model answers A in the first order, and across the four forward orders it picks the gold option 4 times (Figure 2). The 4-bit model picks the gold option in 1 of the four orders. In 4 of the four orders it puts its answer in the same slot, A, whichever option sits there. That pattern is what we call a slot preference: the answer follows the position, and the option content changes. The mean of the four probability vectors has the gold option as its arg max (0.38 against 0.24 for the runner-up). The slot preference cancels when each option has visited each slot once, and the content signal is what remains.

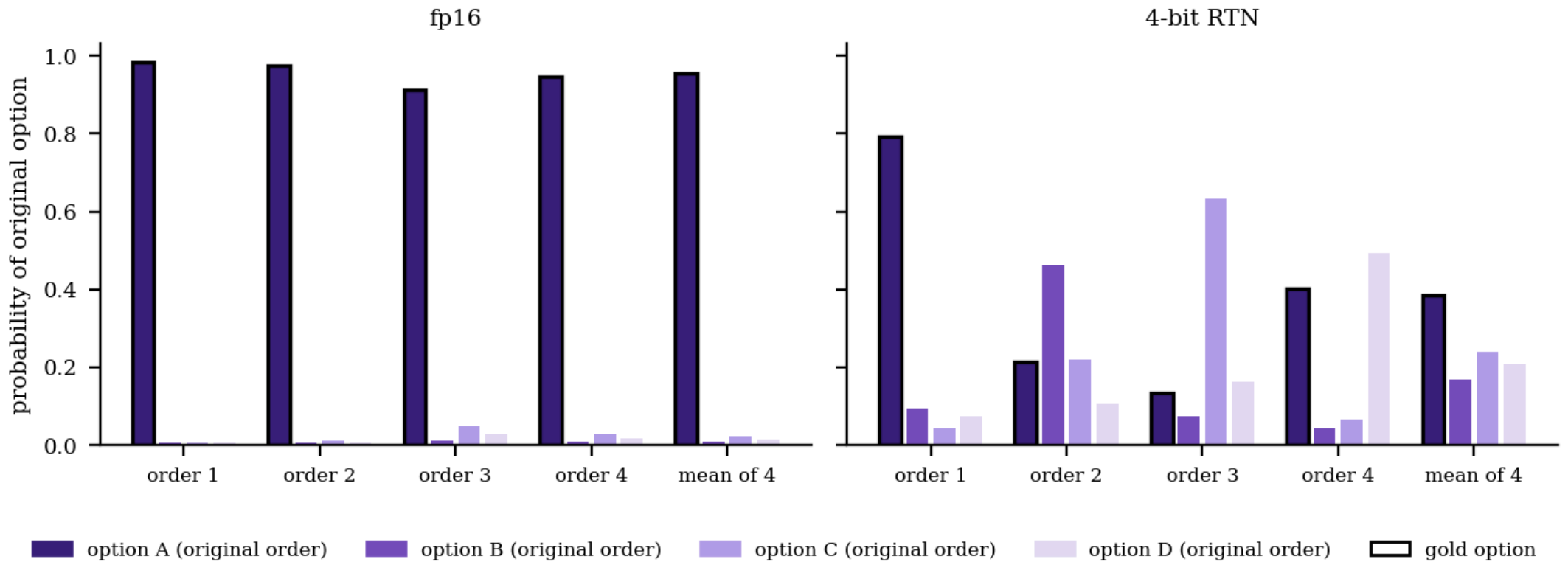

**Figure 2.** One ARC-Easy question under the four forward orders for Qwen3-0.6B. Bars show the probability (renormalised over the four letters) of each original option; the black outline marks the gold option. The right-hand block of each panel is the mean over the four orders.

The example is selected by a rule, not by inspection. It is the first ARC-Easy item (short question) that fp16 gets right under all eight orders, for which the 4-bit model is wrong under at least five orders and its four-order average is right. It illustrates the mechanism; the aggregate numbers that follow show how common it is.

## 4 Accuracy and order sensitivity against weight bits

Figure 3 and Table 2 give the basic picture. Round-to-nearest at 6 bits loses no accuracy against fp16, and 8 bits lose under one point. The accuracy difference is -0.7 (-1.1 to -0.3) (8 bit) and +1.8 (+1.0 to +2.5) (6 bit) points for 0.6B and -0.4 (-0.7 to -0.2) and +0.9 (+0.3 to +1.5) for 1.7B. The 8-bit loss is small, but its interval excludes zero in both models. The 6-bit runs score above fp16 with intervals that exclude zero. We ran no experiment to explain it: rounding may act as a small regulariser, or this may be one of the several comparisons in this paper that clear the interval by chance. The gain does shrink once four orders are averaged: 6-bit minus fp16 at K = 4 is +1.1 (+0.0 to +2.3) and +0.8 (-0.2 to +1.8), and the 6-bit slot skew is below the fp16 one (0.10 against 0.15 and 0.04 against 0.07). Averaging over a full orbit cancels a slot prior, so the gain may be a shifted slot prior and not better content; we did not test this. We therefore read the 6-bit result as "no loss" and do not claim that 6 bits improve a model. The flip rates agree within 1.5 and 2.2 points. Six bits cost 0.39 of an fp16 pass in weight traffic, so on these tasks about 60% of the weight traffic can be dropped at no measurable loss. We use this in the exchange rate of Section 10.

At 4 bits the 0.6B model loses 14.3 accuracy points (-14.3 (-16.1 to -12.5)) and the 1.7B model loses 5.6 (-5.6 (-7.0 to -4.1)). At 3 bits both models are at or just above chance (25.4 (24.9 to 25.8) and 26.6 (25.9 to 27.3) against 25%). The 1.7B model starts higher and degrades less at 4 bits, as expected for a model with more parameters per group of rounded weights, but at 3 bits with this quantizer it has no more room than the small model.

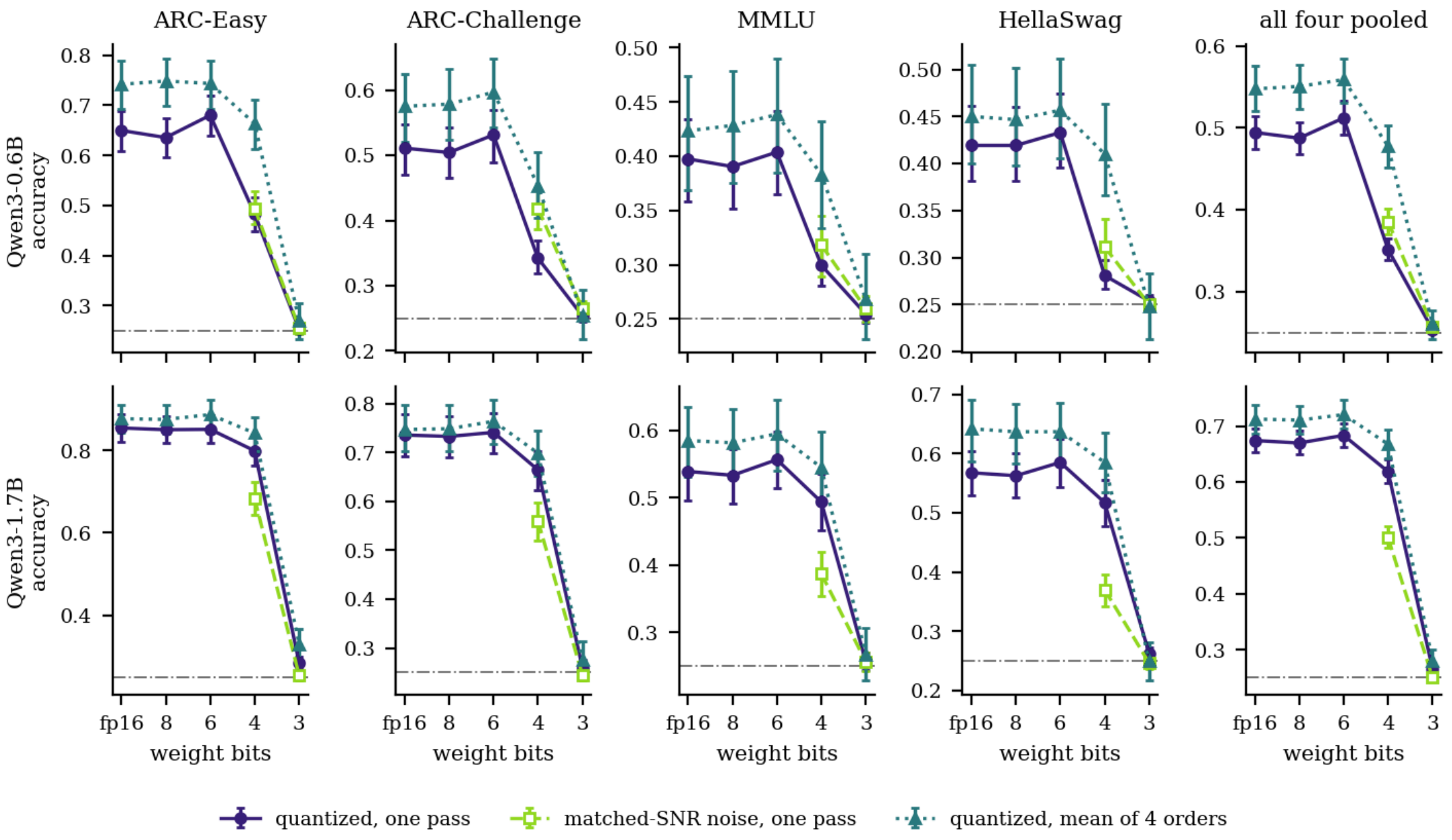

**Figure 3.** One-pass accuracy against weight bits, per task and pooled, for both models. Solid: round-to-nearest. Dashed with open squares: Gaussian noise at the same per-group SNR. Dotted with triangles: the quantized model with the mean probability of four cyclic orders. The grey line is chance. Bars are 95% bootstrap intervals over items.

The flip rate (Figure 4) rises with every lost bit. For 0.6B it goes from 0.42 at fp16 to 0.64 at 4 bits and 0.75 at 3 bits, where it reaches the value of two independent random guesses (0.75). The 1.7B model goes from 0.28 to 0.34 and 0.73. In the fp16 models, 77% (0.6B) and 55% (1.7B) of the questions change answer under at least one of the eight orders, so reordering is routine for small models.

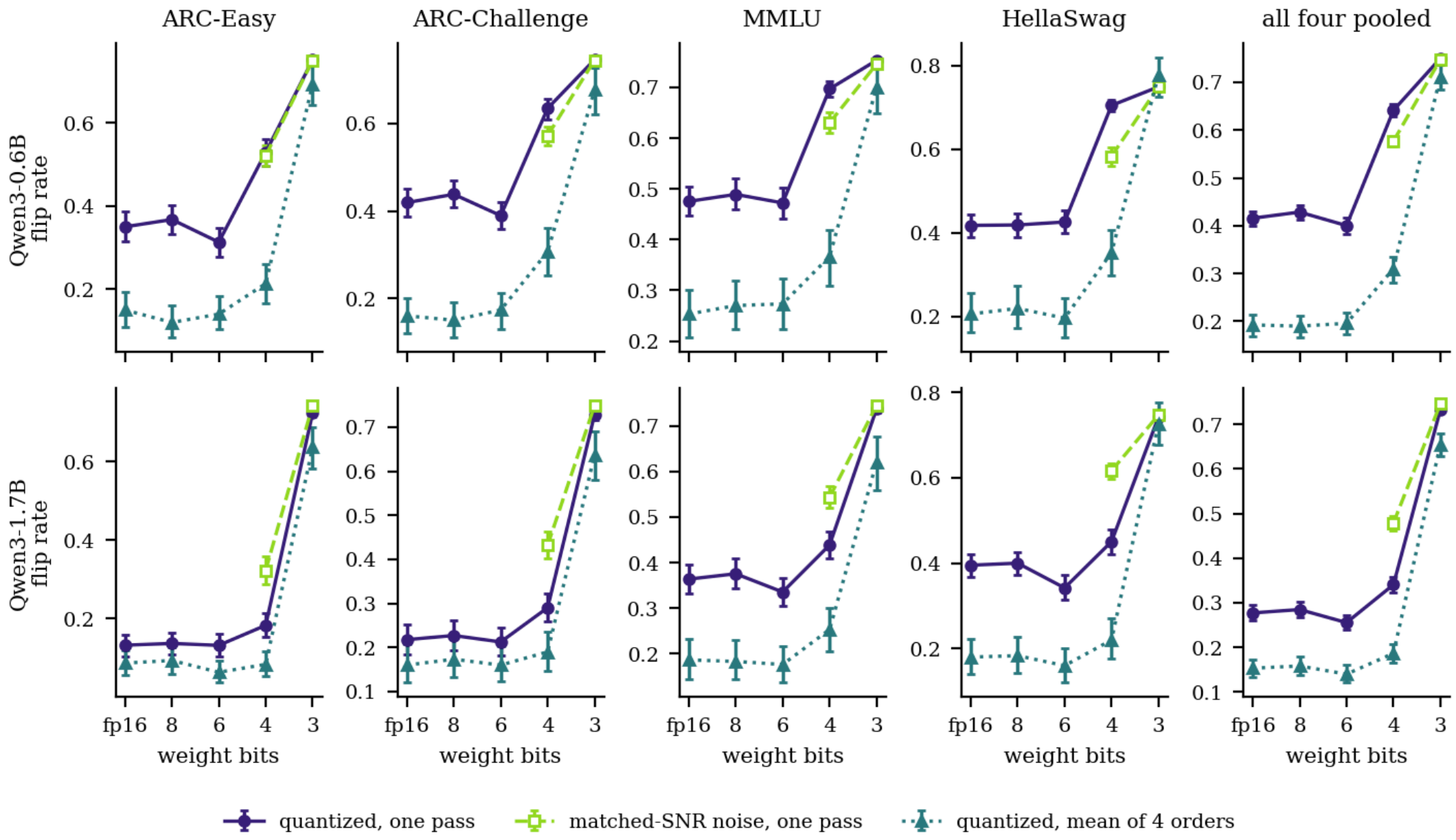

**Figure 4.** Flip rate against weight bits. Solid: one order against one order of the reversed orbit; dotted: after averaging four orders in each orbit (the remaining order sensitivity). Dashed: matched noise. A flip rate of 0.75 is what two independent uniform guesses produce.

Averaging four orders raises accuracy and lowers the flip rate wherever the model still has a content signal, and it does so more for the models that quantization has hurt. The recoverable gap (accuracy of the four-order mean minus one-order accuracy) is +5.3 points at fp16, +12.6 at 4 bits for 0.6B and +3.8 and +4.9 for 1.7B. The interval of the 4-bit gap is +12.6 (+10.5 to +14.7) points for 0.6B and +4.9 (+3.8 to +6.1) for 1.7B. The residual flip rate after averaging (dotted line in Figure 4) still grows with fewer bits: 0.19 to 0.31 for 0.6B and 0.15 to 0.19 for 1.7B. Averaging over cyclic orders therefore removes only part of the extra sensitivity of the quantized models.

**Table 2.** Main results, four tasks pooled (1,200 items per model). Accuracy in percent (95% CI) for one order and for the mean probability of four orders; flip rates are shares of order pairs; 'any flip' is the share of questions whose answer is not identical under all eight orders; slot skew is the total variation distance of the chosen-slot distribution from uniform; letter mass is the share of next-token probability that falls on the four letters.

| Model | Configuration | Accuracy, one order (%) | Accuracy, mean of 4 orders (%) | Flip rate, one order | Flip rate after 4 orders | Any flip | Slot skew | Letter mass |
|---|---|---|---|---|---|---|---|---|
| 0.6B | fp16 (reference) | 49.4 (47.4 to 51.5) | 54.8 (52.1 to 57.5) | 0.416 (0.400 to 0.431) | 0.193 (0.168 to 0.215) | 76.9% | 0.146 | 78.7% |
| 0.6B | 8-bit RTN | 48.7 (46.8 to 50.7) | 55.0 (52.3 to 57.7) | 0.428 (0.412 to 0.444) | 0.190 (0.166 to 0.212) | 78.8% | 0.165 | 78.7% |
| 0.6B | 6-bit RTN | 51.2 (49.2 to 53.2) | 55.9 (53.1 to 58.5) | 0.400 (0.383 to 0.417) | 0.196 (0.173 to 0.218) | 73.9% | 0.096 | 82.0% |
| 0.6B | 4-bit RTN | 35.1 (33.8 to 36.5) | 47.8 (45.1 to 50.3) | 0.641 (0.629 to 0.654) | 0.310 (0.282 to 0.336) | 92.9% | 0.526 | 56.7% |
| 0.6B | 3-bit RTN | 25.4 (24.9 to 25.8) | 26.0 (24.2 to 27.8) | 0.750 (0.747 to 0.752) | 0.712 (0.686 to 0.736) | 100.0% | 0.624 | 35.3% |

| Model | Configuration | Accuracy, one order (%) | Accuracy, mean of 4 orders (%) | Flip rate, one order | Flip rate after 4 orders | Any flip | Slot skew | Letter mass |
|---|---|---|---|---|---|---|---|---|
| 0.6B | noise at 4-bit SNR | 38.6 (37.0 to 40.1) | 45.4 (42.9 to 48.0) | 0.577 (0.565 to 0.588) | 0.307 (0.281 to 0.333) | 95.1% | 0.147 | 93.5% |
| 0.6B | noise at 3-bit SNR | 25.7 (25.2 to 26.3) | 26.0 (24.2 to 27.8) | 0.748 (0.745 to 0.751) | 0.708 (0.683 to 0.731) | 100.0% | 0.451 | 4.5% |
| 0.6B | noise at 4-bit SNR, draw 2 | 33.4 (32.2 to 34.7) | 44.5 (41.9 to 47.1) | 0.658 (0.647 to 0.667) | 0.323 (0.295 to 0.349) | 96.7% | 0.372 | 44.3% |
| 0.6B | noise at 3-bit SNR, draw 2 | 24.8 (24.1 to 25.6) | 25.6 (23.9 to 27.3) | 0.742 (0.738 to 0.747) | 0.736 (0.711 to 0.760) | 100.0% | 0.239 | 13.6% |
| 1.7B | fp16 (reference) | 67.4 (65.3 to 69.6) | 71.2 (68.7 to 73.8) | 0.277 (0.261 to 0.295) | 0.153 (0.133 to 0.173) | 54.7% | 0.066 | 73.2% |
| 1.7B | 8-bit RTN | 67.0 (64.9 to 69.1) | 71.0 (68.5 to 73.6) | 0.285 (0.268 to 0.302) | 0.158 (0.138 to 0.180) | 56.1% | 0.067 | 66.1% |
| 1.7B | 6-bit RTN | 68.3 (66.2 to 70.5) | 72.0 (69.5 to 74.6) | 0.256 (0.240 to 0.272) | 0.140 (0.121 to 0.160) | 51.7% | 0.043 | 78.9% |
| 1.7B | 4-bit RTN | 61.8 (59.8 to 64.0) | 66.8 (64.2 to 69.3) | 0.341 (0.324 to 0.358) | 0.187 (0.165 to 0.208) | 64.2% | 0.125 | 82.7% |
| 1.7B | 3-bit RTN | 26.6 (25.9 to 27.3) | 28.0 (26.1 to 30.0) | 0.733 (0.728 to 0.737) | 0.655 (0.628 to 0.680) | 100.0% | 0.471 | 7.6% |
| 1.7B | noise at 4-bit SNR | 50.0 (48.1 to 52.0) | 57.0 (54.2 to 59.6) | 0.479 (0.463 to 0.494) | 0.271 (0.245 to 0.297) | 82.8% | 0.186 | 87.3% |
| 1.7B | noise at 3-bit SNR | 25.1 (24.4 to 25.7) | 24.2 (22.8 to 26.0) | 0.746 (0.743 to 0.749) | 0.719 (0.694 to 0.745) | 100.0% | 0.447 | 2.2% |

Per task, the pattern is the same in every panel of Figure 3. The MMLU and HellaSwag panels of the 0.6B model start close to chance in absolute terms (fp16 39.8% and 41.9%), so a 4-bit loss there is a loss of a small margin over chance. Appendix B lists the task-level numbers for fp16 and 4 bits.

## 5 The answer-slot prior

Each model has a preferred slot even at fp16. The 0.6B model puts 40% of its answers in slot D (uniform would be 25%), and the 1.7B model puts 29% in slot A. The slot skew is 0.15 and 0.07 (Figure 5). This is the selection bias of Zheng et al. (arXiv:2309.03882) in its simplest form, and the cyclic design measures it without any help from the answer key.

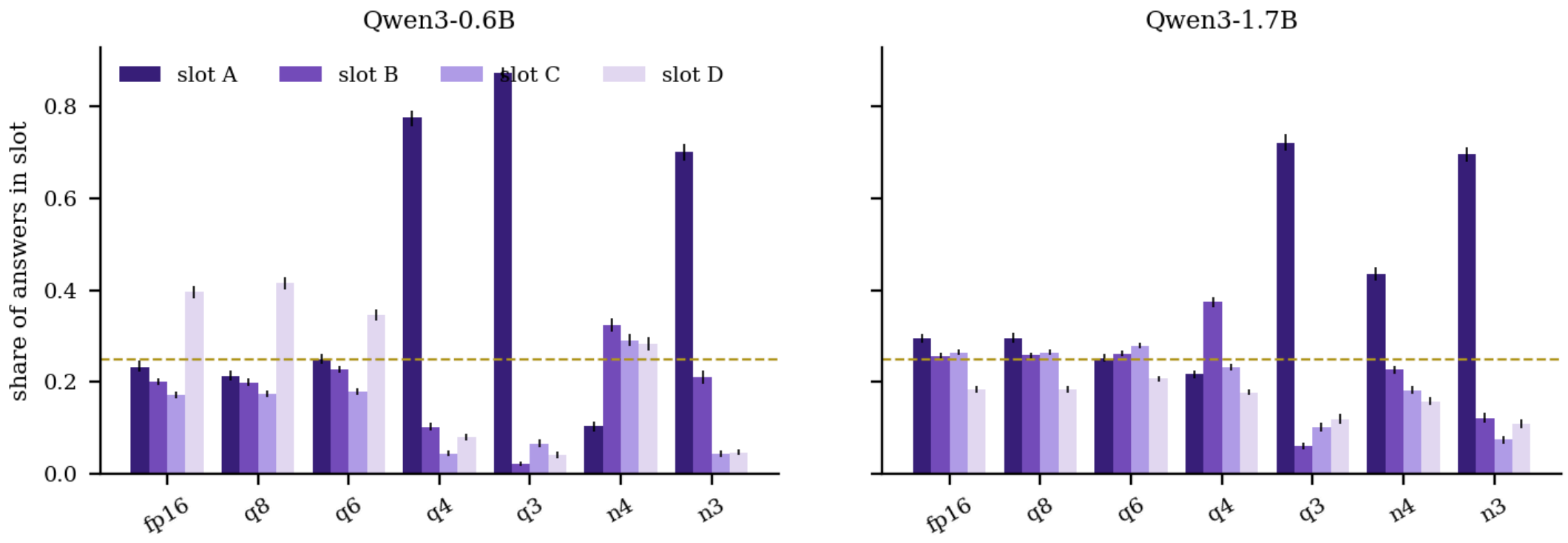

**Figure 5.** Share of answers that land in each letter slot (pooled over tasks and over the eight orders). The red dashed line is the uniform share, 0.25. Bars are 95% bootstrap intervals.

Quantization changes the prior, and by a very different amount in the two models. At 4 bits the 0.6B model answers slot A in 78% of cases and the 1.7B model answers slot B in 37%; the slot skew becomes 0.53 and 0.12. The same-slot share, which is the part of the flip rate that comes from answering by position, goes from 0.24 to 0.70 (0.6B) and from 0.13 to 0.18 (1.7B). At 3 bits the 0.6B model gives slot A 87% of the time and the 1.7B model gives slot A 72% of the time; both have lost their content signal.

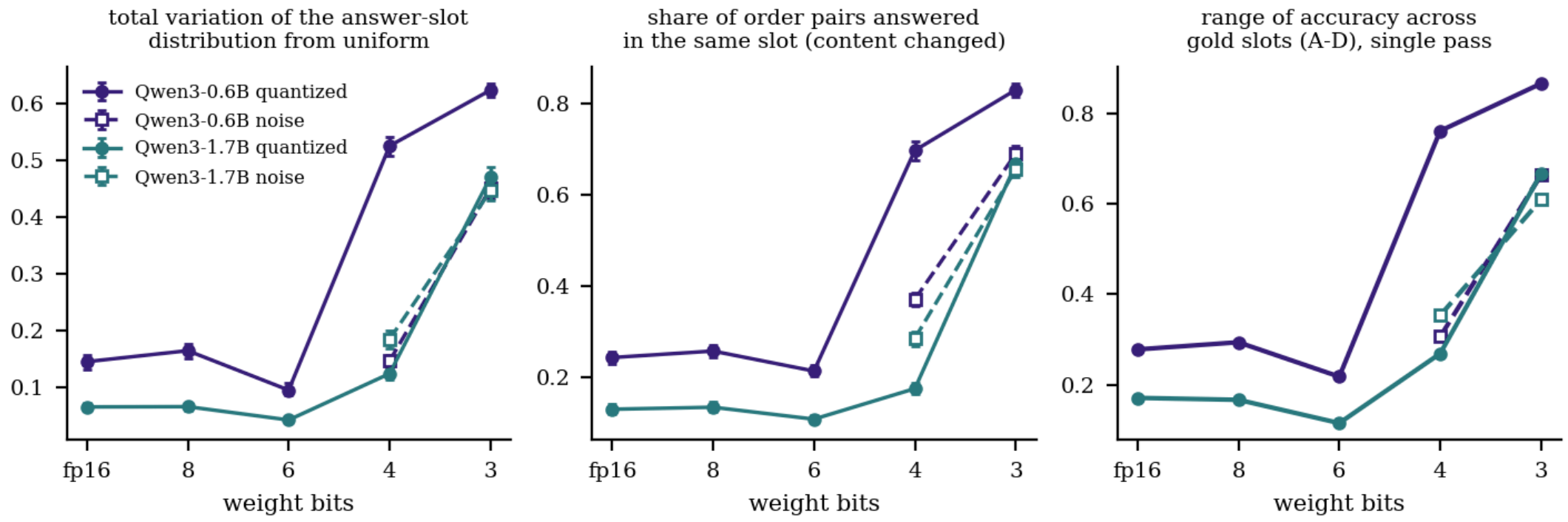

**Figure 6.** Left: total variation distance of the chosen-slot distribution from uniform (slot skew). Middle: share of pairs of orders within an orbit in which the model chose the same slot, which means a different option. Right: range of one-order accuracy across the four gold slots. Dashed lines with open squares are the matched-noise runs.

The accuracy by gold slot (Figure 6, right panel) shows what the prior costs. When the gold option happens to sit in the favoured slot the quantized model is right far more often than average, and elsewhere far less often. For the 4-bit 0.6B model the range across the four slots is 0.76 and for the 4-bit 1.7B model 0.27, against 0.28 and 0.17 at fp16. A benchmark run in the original order, with a gold answer that is roughly uniform over slots, averages these effects. A user whose own questions have a skewed answer position (for example the correct option listed first) would see a very different accuracy for the same weights, and a quantized model would make that gap larger.

Averaging over a full cyclic orbit removes exactly this effect on average, because each option visits each slot once. This is the reason the recoverable gap of Section 4 grows with fewer bits.

## 6 Is it more than accuracy loss? An accuracy-matched baseline and a matched-noise control

### 6.1 An accuracy-matched baseline

A model that gets worse will also flip more, even if nothing about order changes. To ask whether quantization adds order sensitivity beyond that, we need a model of "just worse". Our baseline takes the fp16 answers and replaces each, independently per order, by a uniformly random option with probability r, and sets r so that accuracy equals the quantized model's accuracy: r = (acc(fp16) - acc(config)) / (acc(fp16) - 0.25). The flip rate of this baseline is (1-r)^2 F0 + 0.75 (1 - (1-r)^2), where F0 is the fp16 flip rate, and its slot skew is (1-r) times the fp16 skew. The baseline is crude (real errors are not uniform) but it is a fixed yardstick that uses only fp16 data.

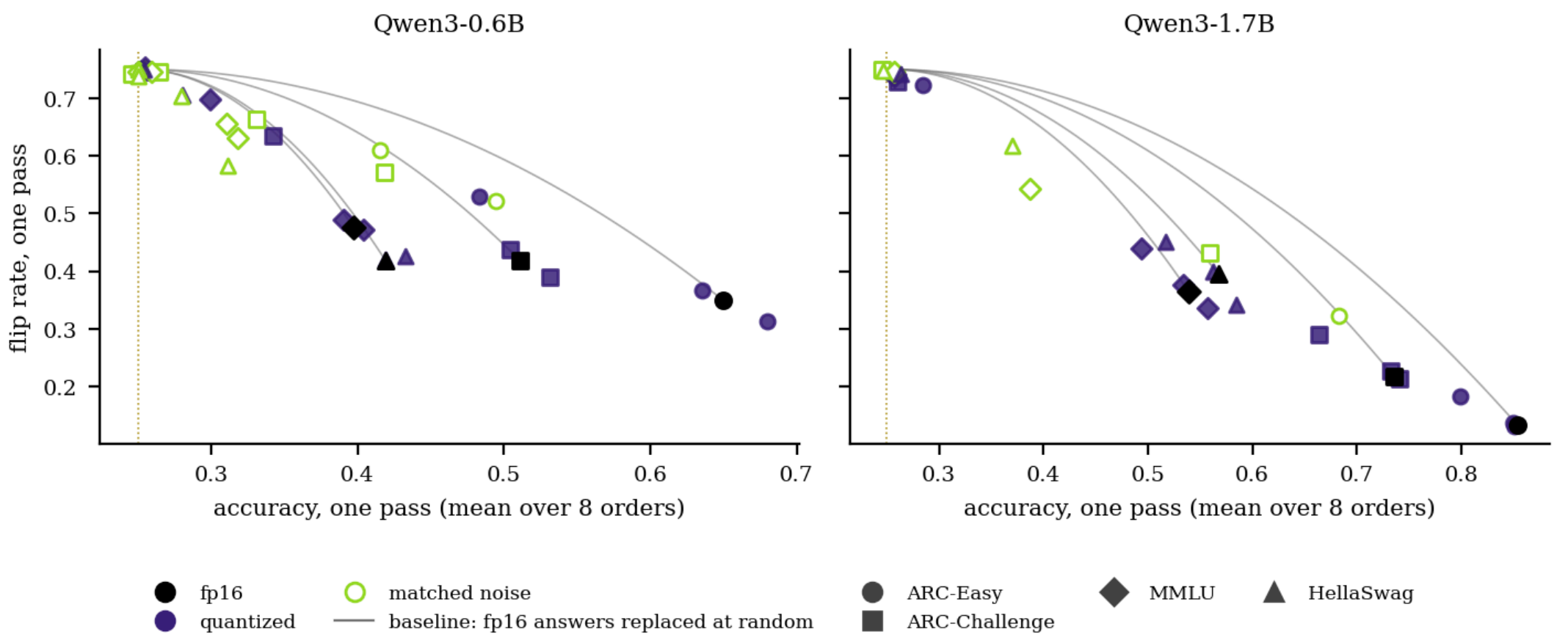

**Figure 7.** Accuracy and flip rate of every configuration, per task. Black: fp16; filled blue: quantized; open orange: matched noise. The grey curves start at the fp16 point of each task and show the flip rate that would result if fp16 answers were replaced by a uniform random option with the probability that matches the observed accuracy.

Table 3 gives the result. At 4 bits the observed flip rate minus the baseline is -5.1 (-6.2 to -4.0) points for 0.6B and -5.3 (-7.5 to -2.9) for 1.7B. The two models give negative excesses (fewer flips than the baseline). Quantization adds order sensitivity in the sense that flips rise with lost bits, but the rise is no larger than the rise that follows from lost accuracy. Real errors are correlated across orders (a model often repeats the same wrong option), which lowers flips relative to independent errors, so a negative excess is weak evidence against extra flips.

For the slot skew, the baseline predicts a value that shrinks as the model is diluted, while the observed skew at 4 bits is 0.53 against a baseline of 0.06 (0.6B) and 0.12 against 0.06 (1.7B). The excess over the baseline is +46.5 (+44.3 to +48.8) points for 0.6B and +6.7 (+5.5 to +7.9) for 1.7B. The slot prior is therefore the part of order sensitivity that accuracy loss does not account for, large in the model that collapsed at 4 bits and smaller in the other.

**Table 3.** Observed order sensitivity against the accuracy-matched baseline. r is the share of fp16 answers that must be replaced by a uniform random option to lose as much accuracy as the configuration does. Excess is in points (100 times the difference of rates), with a 95% bootstrap interval over items.

| Model | Configuration | r | Flip rate | Baseline | Excess (points) | Slot skew | Baseline | Excess (points) |
|---|---|---|---|---|---|---|---|---|
| 0.6B | 8-bit RTN | 0.03 | 0.428 | 0.435 | -0.6 (-1.5 to +0.3) | 0.165 | 0.142 | +2.4 (+1.9 to +2.9) |
| 0.6B | 6-bit RTN | 0.00 | 0.400 | 0.416 | -1.5 (-2.5 to -0.6) | 0.096 | 0.146 | -5.0 (-5.8 to -3.8) |
| 0.6B | 4-bit RTN | 0.59 | 0.641 | 0.692 | -5.1 (-6.2 to -4.0) | 0.526 | 0.061 | +46.5 (+44.3 to +48.8) |
| 0.6B | 3-bit RTN | 0.99 | 0.750 | 0.750 | +0.0 (-0.2 to +0.2) | 0.624 | 0.002 | +62.2 (+61.0 to +63.4) |
| 0.6B | noise at 4-bit SNR | 0.45 | 0.577 | 0.647 | -7.0 (-8.7 to -5.2) | 0.147 | 0.081 | +6.6 (+5.1 to +8.1) |
| 0.6B | noise at 3-bit SNR | 0.97 | 0.748 | 0.750 | -0.2 (-0.5 to +0.1) | 0.451 | 0.004 | +44.7 (+42.8 to +46.6) |
| 1.7B | 8-bit RTN | 0.01 | 0.285 | 0.287 | -0.2 (-0.9 to +0.4) | 0.067 | 0.066 | +0.1 (-0.1 to +0.3) |
| 1.7B | 6-bit RTN | 0.00 | 0.256 | 0.277 | -2.2 (-3.0 to -1.5) | 0.043 | 0.066 | -2.3 (-2.7 to -1.6) |
| 1.7B | 4-bit RTN | 0.13 | 0.341 | 0.393 | -5.3 (-7.5 to -2.9) | 0.125 | 0.058 | +6.7 (+5.5 to +7.9) |

| Model | Configuration | r | Flip rate | Baseline | Excess (points) | Slot skew | Baseline | Excess (points) |
|---|---|---|---|---|---|---|---|---|
| 1.7B | 3-bit RTN | 0.96 | 0.733 | 0.749 | -1.7 (-2.1 to -1.3) | 0.471 | 0.003 | +46.9 (+45.1 to +48.7) |
| 1.7B | noise at 4-bit SNR | 0.41 | 0.479 | 0.586 | -10.7 (-12.3 to -8.9) | 0.186 | 0.039 | +14.6 (+13.1 to +16.1) |
| 1.7B | noise at 3-bit SNR | 1.00 | 0.746 | 0.750 | -0.4 (-0.7 to -0.1) | 0.447 | 0.000 | +44.7 (+42.9 to +46.3) |

## 6.2 What the flips consist of

Figure 8 splits the flipped pairs into those where one of the two answers is correct and those where both are wrong but different. The second part is order sensitivity that accuracy alone cannot see. For the fp16 0.6B model 34% of the flipped pairs are of the both-wrong kind; at 4 bits it is 45%, and for 1.7B 25% and 29%. The share of both-wrong flips grows as accuracy falls, because there are more wrong answers to disagree about. This agrees with the report of Ok et al. (arXiv:2610.20993) that aggregation can lower flips partly by settling on the same wrong option. The reverse also holds: a model that is wrong has many ways to be wrong differently.

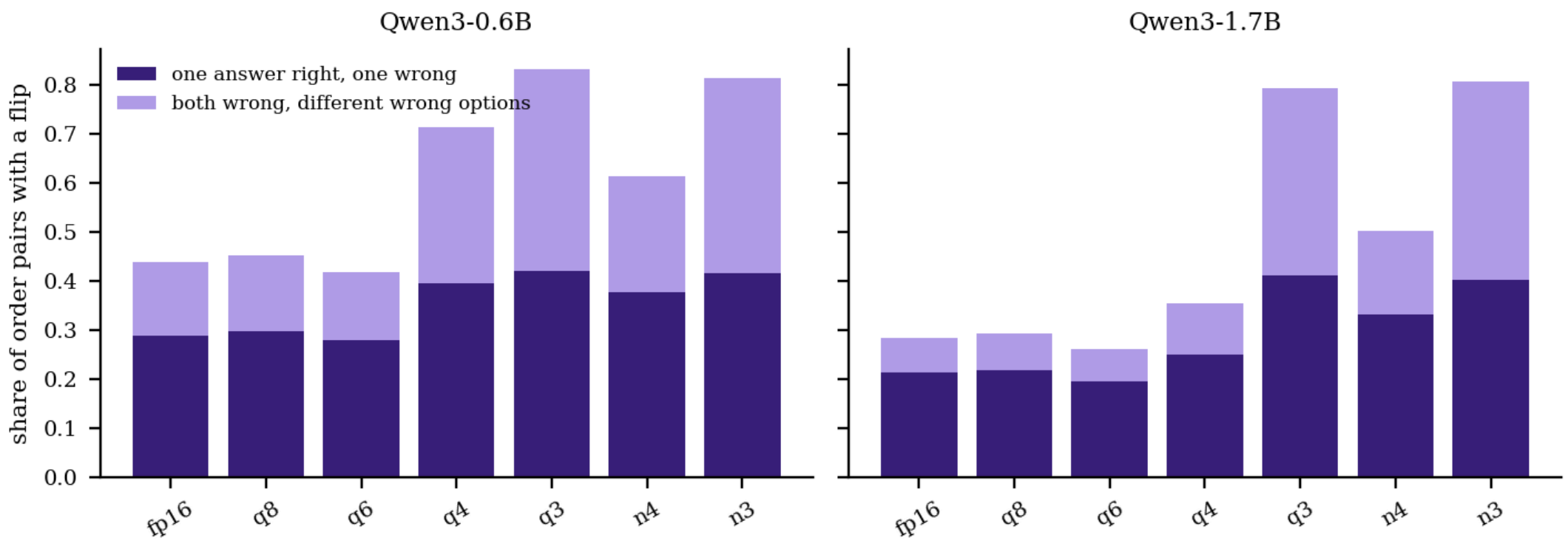

**Figure 8.** Where the flips come from. Each bar is the share of order pairs (all 28 pairs of the eight orders) whose chosen options differ. Dark: one answer correct and the other wrong; light: both wrong, with different wrong options. The second part is invisible to accuracy.

## 6.3 Matched noise

The control asks whether a weight error with the same power but no structure gives the same behaviour (Figure 9). At 4 bits the answer depends on the model. For 0.6B, accuracy is 35.1% under rounding and 38.6% and 33.4% under noise (2 draws); the slot skew is 0.53 under rounding and 0.15 and 0.37 under noise; the one-pass flip rate is 0.64 and 0.58 and 0.66. For 1.7B, accuracy is 61.8% under rounding and 50.0% under noise (1 draw); the slot skew is 0.12 under rounding and 0.19 under noise; the one-pass flip rate is 0.34 and 0.48.

Rounding minus the first noise draw is -3.4 (-5.0 to -1.9) accuracy points for 0.6B ($p<0.001$) and +11.9 (+10.2 to +13.4) for 1.7B ($p<0.001$). For the flip rate the differences are +6.5 (+5.2 to +7.9) and -13.8 (-15.5 to -12.2) points. In the 1.7B model rounding is gentler than the pooled noise draw (on ARC-Easy, where a second 1.7B draw exists, the two noise draws score 68.3% and 68.1% against 79.9% for rounding). In the 0.6B model the accuracy cost of rounding lies between that of the two noise draws, and the slot skew differs between rounding and the two noise draws in both directions. A single noise draw is therefore not a safe reference at this noise level: the 0.6B model at 4-bit SNR sits near the edge where its answers turn into a slot rule, and which side a draw lands on varies. The second noise draw at 4 bits gives, against the first, 0.6B: slot skew 0.15 and 0.37, accuracy 38.6% and 33.4%. At 3 bits both models are at chance under both perturbations and both show a slot preference there (0.45 and 0.45 for noise), which suggests that a model without a content signal falls back on a slot, whatever broke it.

Our data do not show a slot prior that belongs to rounding. They show that the slot skew grows with the amount of damage and that noise can produce it as well as rounding. They also show that perturbations of equal SNR can behave differently: the accuracy lost at 4 bits is about the same for the two perturbations in the 0.6B model and very different in the 1.7B model. This agrees with the remark of Zhou et al. (arXiv:2608.08188) that degradation is not set by bit-width alone.

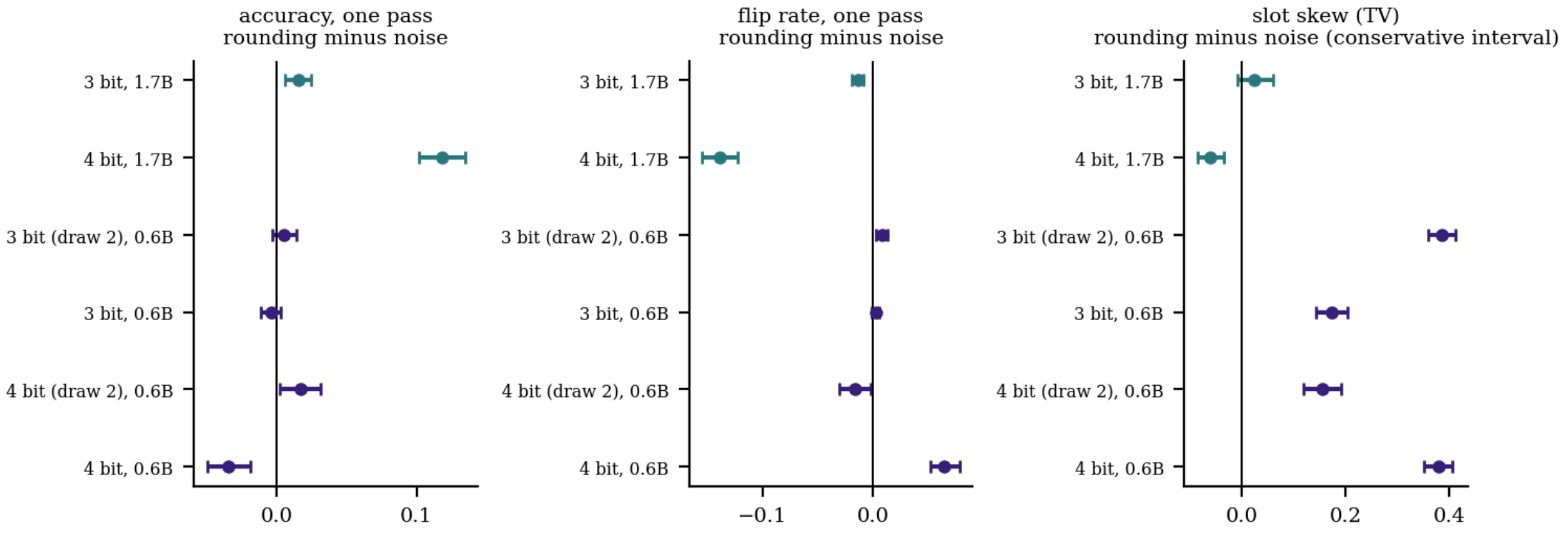

**Figure 9.** Quantized minus matched-noise runs on the same items (paired bootstrap, pooled). Left and middle: one-pass accuracy and flip rate. Right: slot skew, with a conservative interval that combines the two separate intervals.

We found no weight statistic that separates the two perturbations. Total weight energy after rounding is close to that after noise (1.0131 against 1.0123 at 4 bits). Rounding sets 16% of the 4-bit weights to exactly zero and its error is a deterministic function of the weights, which could make it consistent across layers in a way independent noise is not; we did not test this.

## 7 Which questions are fragile

Pezeshkpour and Hruschka (arXiv:2308.11483) conjecture that order sensitivity arises when a model is unsure between its top two or three options, and Ok et al. (arXiv:2610.20993) find that the flip rate tracks output uncertainty. Figure 10 checks this for fp16: in the bin with the smallest margin between the top two options, the 0.6B model flips 0.72 of order pairs, and in the bin with the largest margin 0.00. For 1.7B the values are 0.70 and 0.02. The margin is measured on fp16 itself, so the relation is partly built in (a small margin means that a small push changes the answer), but it shows that the confident questions are stable and the uncertain ones are not.

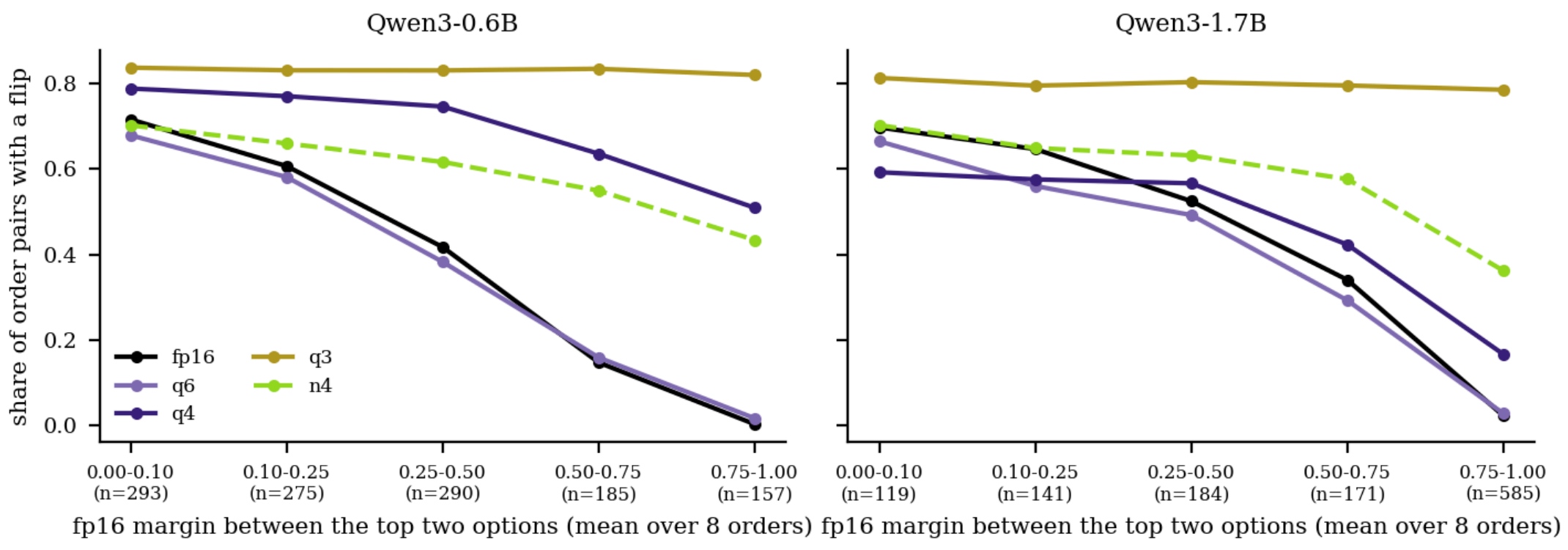

**Figure 10.** Flip rate against the fp16 margin between the two most likely options (the mean-over-orders probability of the best option minus that of the second best). Each point averages the questions in one margin bin; bin sizes are in the axis labels.

Under quantization the relation flattens. For 4-bit weights the smallest and largest margin bins flip 0.79 and 0.51 of order pairs for 0.6B and 0.59 and 0.17 for 1.7B. Questions that the fp16 model answers with a wide margin are not safe once the weights are rounded. For the 0.6B model the slot attractor of Section 5 is a likely reason, which we did not test: when the model leans on one slot, even a question with a clear fp16 answer is answered by position.

Which questions are fragile is therefore not stable across configurations (Table 4). The 4-bit flip share of a question correlates with its fp16 flip share at 0.20 (0.6B) and 0.64 (1.7B), while fp16 against 8-bit gives 0.95 and 0.97. A 4-bit perturbation re-draws which questions are fragile, most in the 0.6B model. The correlation of the fp16 flip share with that of the 4-bit matched-noise run is 0.34 (0.6B) and 0.50 (1.7B). Two independent noise draws at the same SNR correlate at only 0.15 for 0.6B, no higher than rounding with fp16, so the question ranking is specific to each perturbed copy of the model. For 1.7B there is one full noise draw and no such comparison.

**Table 4.** Do the same questions stay fragile? Spearman correlation (average ranks for ties) of the per-question flip share between two configurations, and the overlap (intersection over union, ties broken at random) of their most fragile quarters. Two independent random quarters would overlap at 0.14.

| Model | Configurations | Spearman | Top-quarter overlap |
|---|---|---|---|
| 0.6B | fp16 (reference) / 8-bit RTN | 0.95 | 0.78 |
| 0.6B | fp16 (reference) / 4-bit RTN | 0.20 | 0.18 |
| 0.6B | fp16 (reference) / noise at 4-bit SNR | 0.34 | 0.27 |
| 0.6B | 4-bit RTN / noise at 4-bit SNR | 0.31 | 0.19 |
| 0.6B | noise at 4-bit SNR / noise at 4-bit SNR, draw 2 | 0.15 | 0.18 |
| 0.6B | fp16 (reference) / noise at 4-bit SNR, draw 2 | 0.15 | 0.17 |
| 0.6B | 4-bit RTN / noise at 4-bit SNR, draw 2 | 0.23 | 0.16 |
| 0.6B | 4-bit RTN / 3-bit RTN | 0.09 | 0.16 |
| 0.6B | fp16 (reference) / 3-bit RTN | 0.03 | 0.14 |
| 1.7B | fp16 (reference) / 8-bit RTN | 0.97 | 0.80 |
| 1.7B | fp16 (reference) / 4-bit RTN | 0.64 | 0.36 |
| 1.7B | fp16 (reference) / noise at 4-bit SNR | 0.50 | 0.33 |
| 1.7B | 4-bit RTN / noise at 4-bit SNR | 0.48 | 0.29 |
| 1.7B | 4-bit RTN / 3-bit RTN | 0.14 | 0.19 |
| 1.7B | fp16 (reference) / 3-bit RTN | 0.11 | 0.17 |
| 0.6B / 1.7B | fp16 (reference) in both | 0.27 | 0.23 |
| 0.6B / 1.7B | 4-bit RTN in both | 0.20 | 0.17 |

A question's fragility under one weight setting predicts little about its fragility under another for the 0.6B model and a moderate share of it for 1.7B, so a fragile-question list built on fp16 is a weak guide to where extra permutations are needed on a quantized model. Because the flip share is computed from the model's own answers, a user can recompute it without labels, and we return to this in Section 9.

## 8 Averaging over orders: how many, and how

Figure 11 shows what each extra order buys. At fp16 the first extra order recovers +3.5 points for 0.6B and +3.2 for 1.7B, and the third and fourth add less. At 4 bits the first extra order recovers +5.3 points for 0.6B and +3.0 for 1.7B, and the curve is still rising at four orders for 0.6B (K = 4: 47.8%, K = 3: 44.5%). At 3 bits nothing helps the 0.6B model: its four-order accuracy is 26.0%.

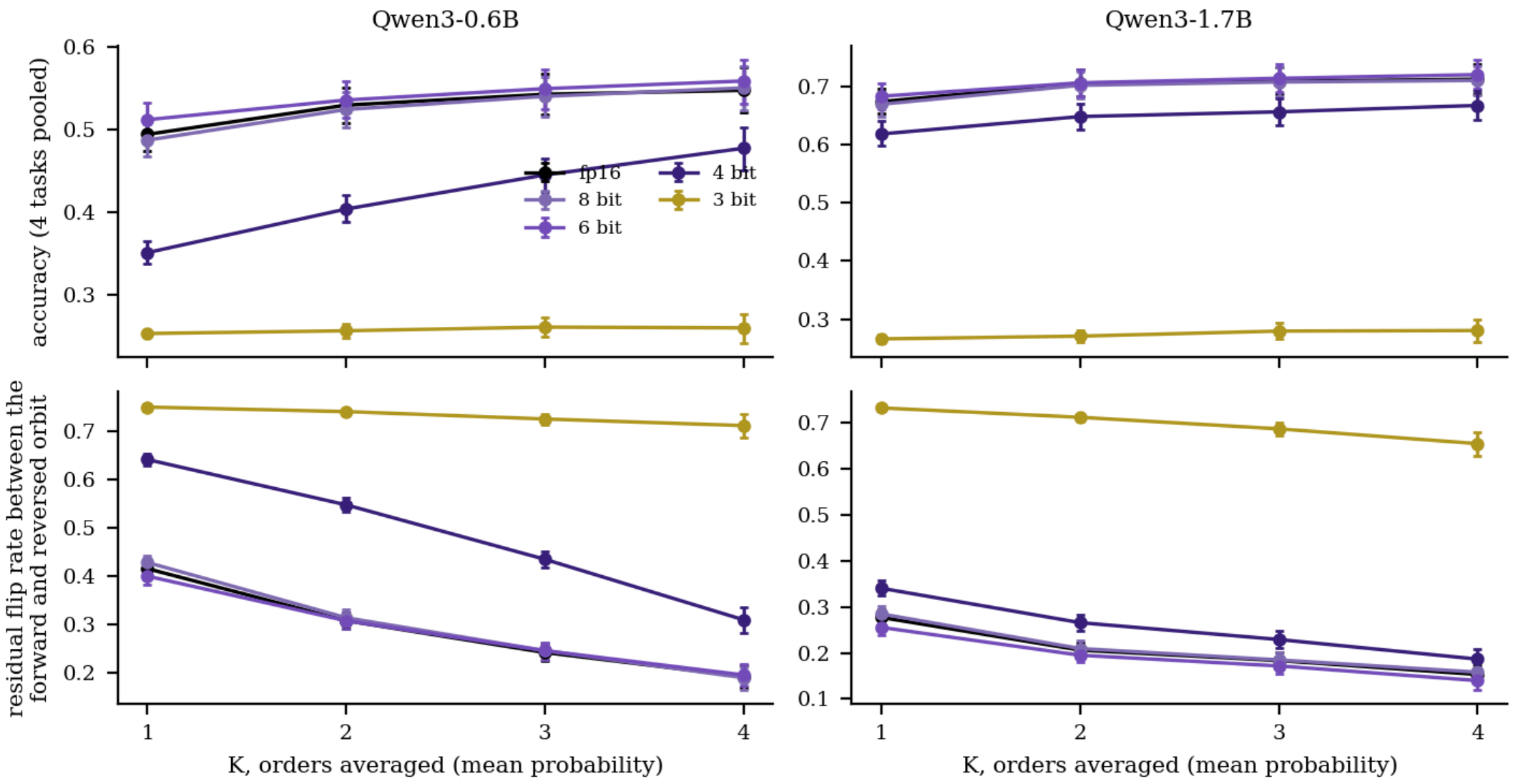

**Figure 11.** Accuracy (top) and residual flip rate (bottom) against the number of averaged orders K, for each weight precision, using the mean probability. Each point averages over all choices of K orders in each orbit; bars are 95% bootstrap intervals (pooled over tasks).

The three aggregation rules (Figure 12) are close in accuracy. At K = 4 the difference between mean log-probability and majority vote at 4 bits is +0.5 (-0.8 to +1.7) points for 0.6B and +0.2 (-0.8 to +1.1) for 1.7B. Between mean log-probability and mean probability the difference is +0.1 (-1.0 to +1.2) and +0.0 (-0.9 to +0.8). The residual flip rate differs more: for 4-bit 0.6B it is 0.310 with the mean probability, 0.271 with the mean log-probability and 0.327 with the vote. The vote discards the confidence of each order, so it keeps more disagreement between orbits. The mean log-probability rewards agreement among orders in the tails and gives the lowest flip rate of the three in 8 of the 8 non-collapsed model and precision settings shown. With only two or four orders none of the rules is clearly the best for accuracy. We use the mean probability in the rest of the paper because it is the simplest and the one whose confidence we study in Section 9.

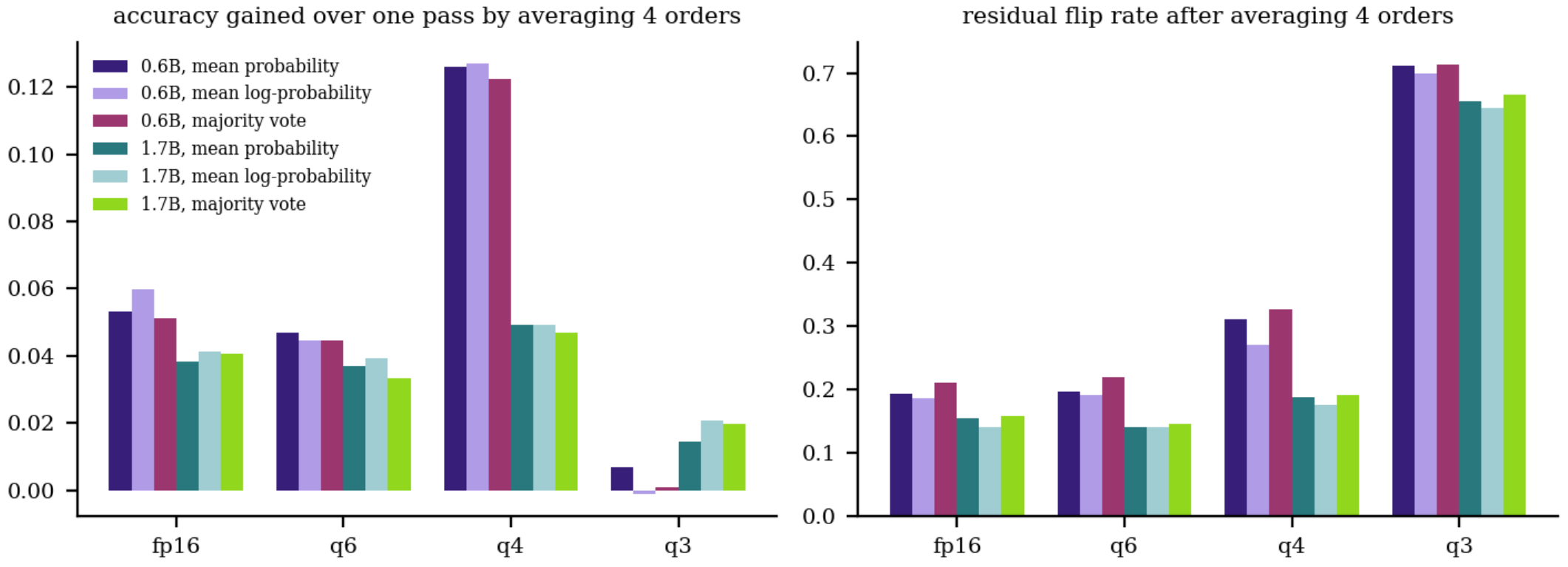

**Figure 12.** Three ways to combine four orders: mean probability, mean log-probability and majority vote. Left: accuracy gained over a single order. Right: residual flip rate. Pooled over tasks.

K = 2 uses two of the four shifts and therefore does not equalise slots, so part of the slot prior stays; K = 4 equalises them exactly. A model that answers by slot (the 4-bit 0.6B model) benefits from every additional order, because each new order moves the favoured slot onto a different option and the sum cancels it. A model with a content signal and a small prior (fp16) gains less because it had less to cancel.

## 9 Confidence and calibration

Small models read through letter probabilities are overconfident, and quantization makes it worse. At fp16 the 0.6B model has a mean top probability of 0.68 at an accuracy of 0.49, an expected calibration error (ECE) of 0.182; at 4 bits the ECE is 0.332. For 1.7B the values are 0.240 and 0.284. This agrees in direction with Proskurina et al. (arXiv:2405.00632), who report a loss of confidence on true labels after 4-bit GPTQ; here the issue is that confidence in the chosen answer stays while accuracy drops.

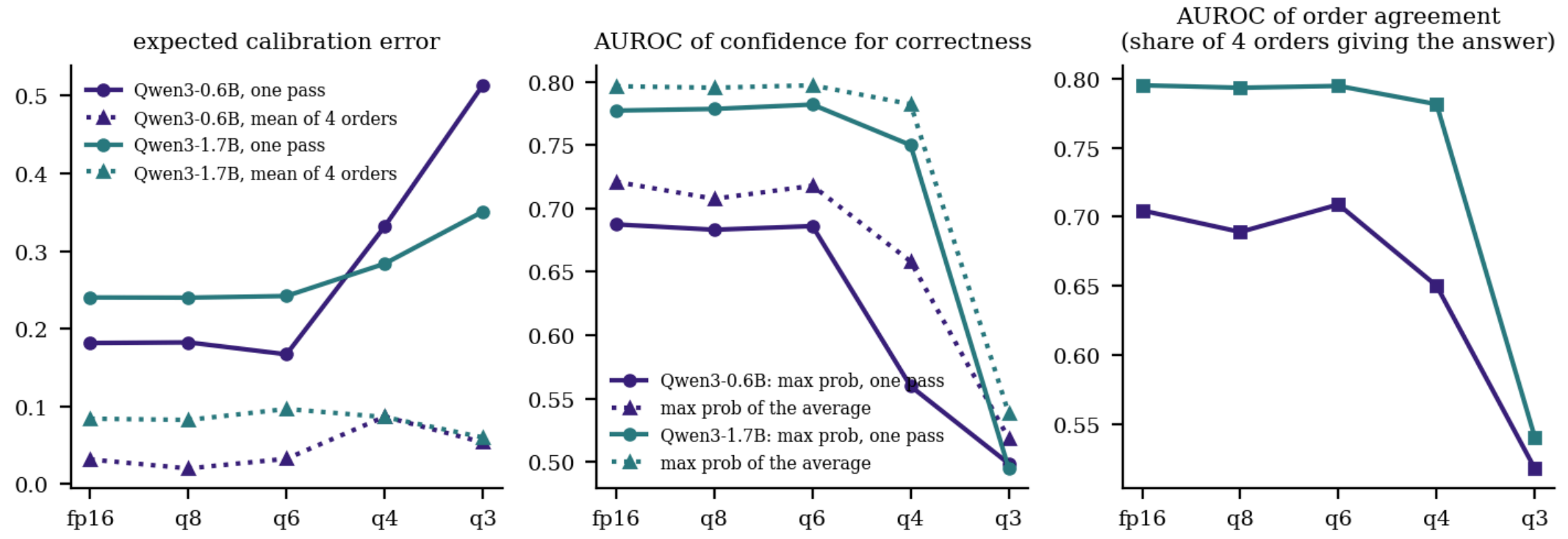

**Figure 13.** Confidence quality of the single-order and four-order answers. Left: expected calibration error (10 equal-width bins; single order pools all eight orders). Middle: AUROC of the top probability for telling correct from wrong answers. Right: AUROC of order agreement for the four-order answer.

Averaging over four orders is a calibration tool as well as an accuracy tool (Table 5). The ECE of the four-order mean is 0.032 at fp16 and 0.087 at 4 bits for 0.6B, and 0.084 and 0.086 for 1.7B, a large fall from the single-order values above. When the orders disagree, the averaged probability of any one option drops, and that disagreement was what the single-order confidence failed to show. Order agreement itself is a label-free signal of correctness: its AUROC for the four-order answer is 0.70 (0.6B, fp16) and 0.80 (1.7B, fp16), against 0.69 and 0.78 for the single-order top probability. At 4 bits the single-order top probability falls to 0.56 (0.6B) and 0.75 (1.7B), close to uninformative for 0.6B, while order agreement keeps 0.65 and 0.78.

**Table 5.** Calibration and confidence ranking, pooled over tasks. Confidence is the top probability among the four letters, for one order (all eight orders pooled) or for the mean probability of the four orders of an orbit. AUROC is the chance that a correct answer gets a higher confidence than a wrong one; 'agreement' is the share of the four orders whose answer equals the four-order answer.

| Model | Configuration | Conf. 1 order | Acc. 1 order | ECE 1 order | Conf. mean of 4 | Acc. mean of 4 | ECE mean of 4 | AUROC conf. 1 | AUROC conf. 4 | AUROC agreement |
|---|---|---|---|---|---|---|---|---|---|---|
| 0.6B | fp16 (reference) | 0.676 | 0.494 | 0.182 | 0.579 | 0.547 | 0.032 | 0.687 | 0.721 | 0.705 |
| 0.6B | 8-bit RTN | 0.670 | 0.487 | 0.183 | 0.570 | 0.550 | 0.020 | 0.683 | 0.708 | 0.689 |
| 0.6B | 6-bit RTN | 0.679 | 0.512 | 0.167 | 0.590 | 0.559 | 0.033 | 0.686 | 0.718 | 0.709 |
| 0.6B | 4-bit RTN | 0.683 | 0.351 | 0.332 | 0.391 | 0.477 | 0.087 | 0.560 | 0.658 | 0.650 |
| 0.6B | 3-bit RTN | 0.767 | 0.254 | 0.514 | 0.314 | 0.260 | 0.054 | 0.499 | 0.518 | 0.518 |
| 0.6B | noise at 4-bit SNR | 0.609 | 0.386 | 0.224 | 0.455 | 0.454 | 0.022 | 0.612 | 0.686 | 0.675 |
| 1.7B | fp16 (reference) | 0.914 | 0.674 | 0.240 | 0.797 | 0.713 | 0.084 | 0.777 | 0.796 | 0.795 |
| 1.7B | 8-bit RTN | 0.910 | 0.670 | 0.240 | 0.791 | 0.710 | 0.083 | 0.778 | 0.795 | 0.793 |
| 1.7B | 6-bit RTN | 0.926 | 0.683 | 0.242 | 0.817 | 0.720 | 0.097 | 0.782 | 0.797 | 0.795 |
| 1.7B | 4-bit RTN | 0.902 | 0.618 | 0.284 | 0.748 | 0.667 | 0.086 | 0.750 | 0.782 | 0.782 |
| 1.7B | 3-bit RTN | 0.617 | 0.266 | 0.351 | 0.341 | 0.280 | 0.060 | 0.495 | 0.538 | 0.540 |
| 1.7B | noise at 4-bit SNR | 0.819 | 0.500 | 0.319 | 0.622 | 0.570 | 0.057 | 0.692 | 0.731 | 0.724 |

On a quantized model the single-pass confidence is a poor guide to when an answer can be trusted, and the disagreement among a few reordered passes is a better one, at the price of the extra passes. Our experiments test only the ranking quality of these signals, not thresholds or selective-answering policies.

# 10 An exchange rate between bits and permutations

Section 4 showed that averaging over orders buys more accuracy as the weights get coarser, but each extra pass costs as much as the first, so the trade needs a common price. This section prices both in one unit. A forward pass at b bits per weight reads (b + 0.25) / 16 of the bytes an fp16 pass reads (the 0.25 is the fp16 scale per group of 64). K orders need K passes, so a cell (b, K) costs K (b + 0.25) / 16. We state costs in these units and make no claim about seconds. The unit counts every weight at the stated width; Section 14 shows how much the 6-bit pass cost changes when the fp16 embedding matrix is counted.

## 10.1 The accuracy-cost frontier

Figure 14 plots all 20 cells for each model. The cells on the frontier, from cheap to expensive, are (3 bit, K = 1), (4 bit, K = 1), (6 bit, K = 1), (6 bit, K = 2), (6 bit, K = 3), (6 bit, K = 4) for 0.6B. For 1.7B they are (3 bit, K = 1), (4 bit, K = 1), (6 bit, K = 1), (6 bit, K = 2), (6 bit, K = 3), (6 bit, K = 4). Neither fp16 nor 8-bit weights appear on it: a 6-bit cell with the same K costs less and is no less accurate on the point estimate (its paired advantage over fp16 excludes zero in 1 of 6 cases at K = 2 to 4, so the frontier mostly records that 6 bits cost less than 16 at similar accuracy). Cells with fewer bits and more orders lie below the frontier: 4 bits with K = 4 costs 1.06 and reaches 47.8% (0.6B) and 66.8% (1.7B), while 6 bits with K = 2 costs 0.78 and reaches 53.6% and 70.6%.

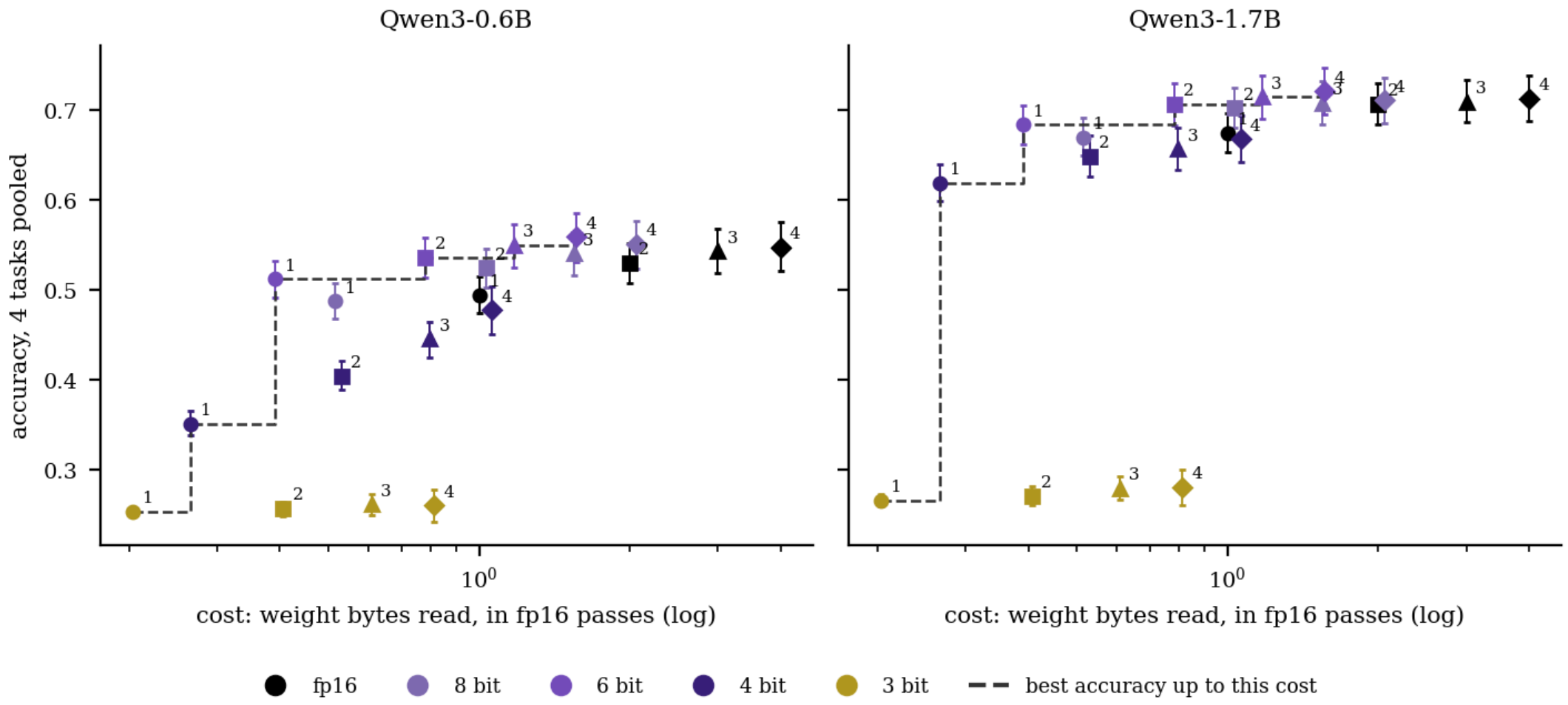

**Figure 14.** Accuracy against cost for every (bits, K) cell, pooled over the four tasks. Each point is a bit width (colour) with K orders averaged (label and marker); cost is the weight bytes read in units of one fp16 pass, on a log scale. The dashed step line is the best accuracy reachable up to a given cost. Bars are 95% bootstrap intervals.

## 10.2 Head-to-head at equal cost

The test of "quantize harder and average" against "keep more bits and answer once" compares a one-order cell with a many-order cell of fewer bits that costs at most 13% more (and at least half as much). Table 6 lists every such pair in our grid. Extra orders win on accuracy for 0.6B in 8 bit x 2 against fp16 x 1 (+3.0 points); 6 bit x 2 against fp16 x 1 (+4.1 points). For 1.7B they win in 8 bit x 2 against fp16 x 1 (+2.8 points); 6 bit x 2 against fp16 x 1 (+3.2 points). More bits win for 0.6B in 3 bit x 2 against 6 bit x 1 (-25.5 points); 4 bit x 2 against 8 bit x 1 (-8.3 points); 3 bit x 2 against 8 bit x 1 (-23.0 points); 4 bit x 2 against fp16 x 1 (-9.0 points); 4 bit x 3 against fp16 x 1 (-4.9 points); 3 bit x 3 against fp16 x 1 (-23.3 points); 3 bit x 4 against fp16 x 1 (-23.4 points). For 1.7B they win in 3 bit x 2 against 6 bit x 1 (-41.3 points); 4 bit x 2 against 8 bit x 1 (-2.1 points); 3 bit x 2 against 8 bit x 1 (-39.9 points); 4 bit x 2 against fp16 x 1 (-2.6 points); 4 bit x 3 against fp16 x 1 (-1.8 points); 3 bit x 3 against fp16 x 1 (-39.5 points); 3 bit x 4 against fp16 x 1 (-39.4 points). The rest are ties within the interval. The pairs are not all at equal cost: the many-order cell costs between 0.53 and 1.06 of the one-order cell, and the pairs share cells, so the counts are not independent. In the 4 pairs per model within 5% below and 7% above equal cost, extra orders win in 1 and 1 (0.6B, 1.7B), more bits win in 2 and 2, and 1 and 1 are ties. On flip rate the fewer-bit, many-order side is lower in 3 of 10 pairs for 0.6B and 5 of 10 for 1.7B.

**Table 6.** Equal-cost comparisons: a many-order cell with fewer bits that costs at most 13% more (and at least half as much) than a one-order cell with more bits. Cost in fp16 passes, many-order cell first. Differences are many-order minus one-order, in points, with 95% paired bootstrap intervals and p values.

| Model | Fewer bits, K orders | More bits, one order | Cost | Accuracy (%) | Accuracy diff. (points) | p | Flip rate | Flip-rate diff. (points) |
|---|---|---|---|---|---|---|---|---|
| 0.6B | 3 bit, K = 2 | 6 bit, K = 1 | 0.41 / 0.39 | 25.7 / 51.2 | -25.5 (-27.7 to -23.1) | p<0.001 | 0.741 / 0.400 | +34.0 (+32.3 to +35.9) |
| 0.6B | 4 bit, K = 2 | 8 bit, K = 1 | 0.53 / 0.52 | 40.4 / 48.7 | -8.3 (-10.0 to -6.6) | p<0.001 | 0.548 / 0.428 | +11.9 (+10.0 to +13.7) |
| 0.6B | 3 bit, K = 2 | 8 bit, K = 1 | 0.41 / 0.52 | 25.7 / 48.7 | -23.0 (-25.2 to -20.9) | p<0.001 | 0.741 / 0.428 | +31.2 (+29.6 to +33.0) |
| 0.6B | 8 bit, K = 2 | fp16, K = 1 | 1.03 / 1.00 | 52.4 / 49.4 | +3.0 (+2.3 to +3.8) | p<0.001 | 0.314 / 0.416 | -10.1 (-11.0 to -9.2) |
| 0.6B | 6 bit, K = 2 | fp16, K = 1 | 0.78 / 1.00 | 53.6 / 49.4 | +4.1 (+3.2 to +5.1) | p<0.001 | 0.308 / 0.416 | -10.8 (-11.9 to -9.6) |
| 0.6B | 4 bit, K = 2 | fp16, K = 1 | 0.53 / 1.00 | 40.4 / 49.4 | -9.0 (-10.6 to -7.4) | p<0.001 | 0.548 / 0.416 | +13.2 (+11.4 to +15.0) |
| 0.6B | 4 bit, K = 3 | fp16, K = 1 | 0.80 / 1.00 | 44.5 / 49.4 | -4.9 (-6.6 to -3.2) | p<0.001 | 0.435 / 0.416 | +1.9 (+0.1 to +3.7) |
| 0.6B | 4 bit, K = 4 | fp16, K = 1 | 1.06 / 1.00 | 47.8 / 49.4 | -1.7 (-3.8 to +0.5) | p=0.128 | 0.310 / 0.416 | -10.6 (-13.3 to -7.9) |
| 0.6B | 3 bit, K = 3 | fp16, K = 1 | 0.61 / 1.00 | 26.1 / 49.4 | -23.3 (-25.5 to -21.0) | p<0.001 | 0.725 / 0.416 | +31.0 (+29.1 to +32.9) |
| 0.6B | 3 bit, K = 4 | fp16, K = 1 | 0.81 / 1.00 | 26.0 / 49.4 | -23.4 (-26.0 to -20.8) | p<0.001 | 0.712 / 0.416 | +29.6 (+26.7 to +32.5) |
| 1.7B | 3 bit, K = 2 | 6 bit, K = 1 | 0.41 / 0.39 | 27.1 / 68.3 | -41.3 (-43.5 to -38.9) | p<0.001 | 0.712 / 0.256 | +45.7 (+43.9 to +47.3) |
| 1.7B | 4 bit, K = 2 | 8 bit, K = 1 | 0.53 / 0.52 | 64.8 / 67.0 | -2.1 (-3.6 to -0.7) | p=0.004 | 0.266 / 0.285 | -1.9 (-3.5 to -0.3) |
| 1.7B | 3 bit, K = 2 | 8 bit, K = 1 | 0.41 / 0.52 | 27.1 / 67.0 | -39.9 (-42.2 to -37.6) | p<0.001 | 0.712 / 0.285 | +42.7 (+40.9 to +44.5) |
| 1.7B | 8 bit, K = 2 | fp16, K = 1 | 1.03 / 1.00 | 70.2 / 67.4 | +2.8 (+2.2 to +3.4) | p<0.001 | 0.210 / 0.277 | -6.8 (-7.6 to -5.9) |
| 1.7B | 6 bit, K = 2 | fp16, K = 1 | 0.78 / 1.00 | 70.6 / 67.4 | +3.2 (+2.5 to +4.0) | p<0.001 | 0.195 / 0.277 | -8.2 (-9.2 to -7.2) |
| 1.7B | 4 bit, K = 2 | fp16, K = 1 | 0.53 / 1.00 | 64.8 / 67.4 | -2.6 (-4.1 to -1.1) | p=0.002 | 0.266 / 0.277 | -1.1 (-2.7 to +0.3) |
| 1.7B | 4 bit, K = 3 | fp16, K = 1 | 0.80 / 1.00 | 65.6 / 67.4 | -1.8 (-3.3 to -0.3) | p=0.022 | 0.229 / 0.277 | -4.8 (-6.4 to -3.1) |
| 1.7B | 4 bit, K = 4 | fp16, K = 1 | 1.06 / 1.00 | 66.8 / 67.4 | -0.7 (-2.3 to +1.1) | p=0.500 | 0.187 / 0.277 | -9.1 (-11.3 to -6.7) |
| 1.7B | 3 bit, K = 3 | fp16, K = 1 | 0.61 / 1.00 | 27.9 / 67.4 | -39.5 (-41.8 to -36.9) | p<0.001 | 0.687 / 0.277 | +41.0 (+38.9 to +42.9) |
| 1.7B | 3 bit, K = 4 | fp16, K = 1 | 0.81 / 1.00 | 28.0 / 67.4 | -39.4 (-42.2 to -36.5) | p<0.001 | 0.655 / 0.277 | +37.8 (+34.8 to +40.6) |

The pattern follows the bit range of the many-order cell. When it has 6 or 8 bits, rounding loss is negligible and the extra orders are a gain. With 8 bits and K = 2 against fp16 once, the difference is +3.0 (+2.3 to +3.8) points (0.6B) and +2.8 (+2.2 to +3.4) points (1.7B) at 3% higher cost. When it has 4 bits, one pass at more bits is better or equal on accuracy: 4 bits with K = 2 against 8 bits once is -8.3 (-10.0 to -6.6) points and -2.1 (-3.6 to -0.7) points. For 4 bits with K = 4 against fp16 once the difference is -1.7 (-3.8 to +0.5) points and -0.7 (-2.3 to +1.1) points, an interval that includes zero, although its flip rate is lower by 10.6 and 9.1 points. At 3 bits no pair is close: 3 bits with K = 2 against 6 bits once is -25.5 (-27.7 to -23.1) points and -41.3 (-43.5 to -38.9) points.

### 10.3 How many bits is a permutation worth?

Another way to express the exchange rate is the bit equivalent of averaging (Figure 15). Starting from 4-bit weights, K = 2 lifts the 0.6B model to the accuracy of a single pass at about 4.7 bits, and K = 4 to about 5.6 bits; for 1.7B the values are 4.9 and 5.5. Raising 4-bit weights to 5.6 bits costs a factor 1.37 in weight bytes, while four orders cost a factor of 4; for 1.7B the factors are 1.36 and 4. In this simple sense bits are cheaper per point of accuracy than permutations whenever the user can still store more bits, and permutations become the only lever when the weights are already as small as the user can afford. Interpolation between the evaluated bit widths is linear and the 5-bit point was not run, so the equivalents are approximate.

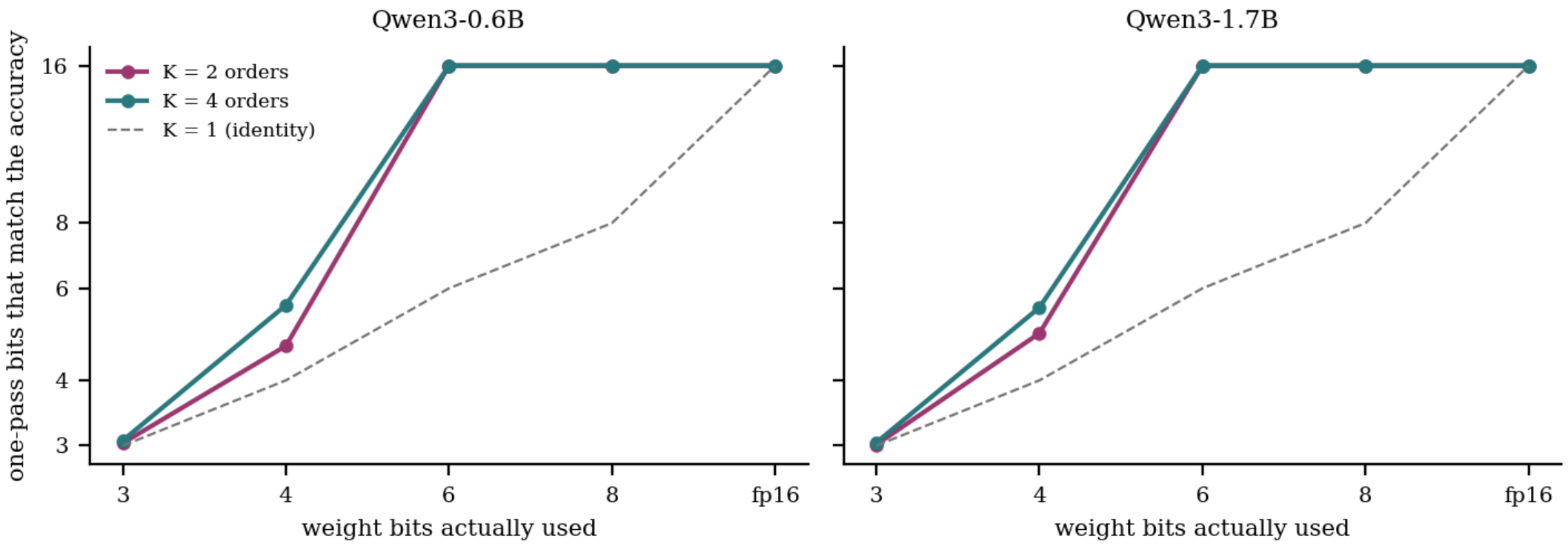

**Figure 15.** Bit equivalent of permutation averaging. For each bit width on the horizontal axis, the vertical axis gives the number of bits at which a single pass would match the accuracy of K orders at that width (piecewise linear between the evaluated widths, capped at 16). A point above the dashed line means that averaging is worth more than a gain in bits.

### 10.4 Per task

The pooled picture hides differences between tasks. Where the fp16 model is well above chance (ARC-Easy for both models) averaging at 4 bits recovers most of the loss against one fp16 pass, but on ARC-Challenge both models stay significantly below it (Table 7). Where it is close to chance (0.6B on MMLU and HellaSwag) there is little for either lever to recover. Table 7 gives two headline comparisons by task.

**Table 7.** Two headline equal-cost comparisons by task (points, 95% paired bootstrap interval; many-order cell minus one-order cell).

| Model | Task | Comparison | Accuracy diff. | Flip-rate diff. |
|---|---|---|---|---|
| 0.6B | ARC-Easy | 6 bit x 2 vs fp16 x 1 | +6.2 (+4.3 to +8.1) | -10.4 (-13.1 to -8.1) |
| 0.6B | ARC-Easy | 4 bit x 4 vs fp16 x 1 | +1.4 (-2.9 to +5.6) | -13.7 (-18.9 to -8.6) |
| 0.6B | ARC-Challenge | 6 bit x 2 vs fp16 x 1 | +5.2 (+3.2 to +7.2) | -13.7 (-15.9 to -11.4) |
| 0.6B | ARC-Challenge | 4 bit x 4 vs fp16 x 1 | -5.8 (-10.0 to -1.4) | -11.2 (-16.5 to -5.7) |
| 0.6B | MMLU | 6 bit x 2 vs fp16 x 1 | +2.8 (+0.8 to +4.5) | -10.1 (-12.5 to -7.9) |
| 0.6B | MMLU | 4 bit x 4 vs fp16 x 1 | -1.4 (-5.8 to +2.6) | -10.9 (-17.0 to -5.0) |
| 0.6B | HellaSwag | 6 bit x 2 vs fp16 x 1 | +2.5 (+1.0 to +4.0) | -8.9 (-10.9 to -6.9) |
| 0.6B | HellaSwag | 4 bit x 4 vs fp16 x 1 | -0.9 (-5.0 to +3.5) | -6.5 (-12.3 to -1.1) |
| 1.7B | ARC-Easy | 6 bit x 2 vs fp16 x 1 | +1.7 (+0.5 to +2.9) | -3.9 (-5.3 to -2.5) |
| 1.7B | ARC-Easy | 4 bit x 4 vs fp16 x 1 | -1.2 (-4.0 to +1.4) | -4.9 (-8.5 to -1.1) |
| 1.7B | ARC-Challenge | 6 bit x 2 vs fp16 x 1 | +1.6 (+0.2 to +2.8) | -5.1 (-6.8 to -3.4) |
| 1.7B | ARC-Challenge | 4 bit x 4 vs fp16 x 1 | -3.7 (-7.2 to -0.3) | -2.8 (-7.1 to +1.6) |

| Model | Task | Comparison | Accuracy diff. | Flip-rate diff. |
|---|---|---|---|---|
| 1.7B | MMLU | 6 bit x 2 vs fp16 x 1 | +4.5 (+2.8 to +6.1) | -8.6 (-11.0 to -6.5) |
| 1.7B | MMLU | 4 bit x 4 vs fp16 x 1 | +0.6 (-3.1 to +4.8) | -11.1 (-16.1 to -5.7) |
| 1.7B | HellaSwag | 6 bit x 2 vs fp16 x 1 | +5.1 (+3.3 to +6.9) | -15.3 (-17.4 to -13.2) |
| 1.7B | HellaSwag | 4 bit x 4 vs fp16 x 1 | +1.8 (-1.8 to +5.2) | -17.5 (-22.0 to -12.7) |

## 11 A decision table

A budget B is the weight traffic one is willing to spend per question, in fp16 passes. Among the cells of the grid that cost at most B, we take the cheapest one whose accuracy is within half a point of the best affordable cell. This keeps cells that differ only by noise (6-bit, 8-bit and fp16 weights) from being told apart by luck. Choosing on the same items that are used to report the result would inflate it, so we cross-fit. The choice is made on one half of the items and scored on the other half, and the two halves swap roles (Table 8).

**Table 8.** If your budget is B (fp16 passes per question), choose these bits and K. 'Chosen' is the cheapest affordable cell whose accuracy is within half a point of the best affordable cell (chosen on all items); accuracy is held out (selected on one half of the items, scored on the other half). 'Best one order' is the best affordable cell with K = 1. The last column says whether the two halves chose the same cell.

| Model | Budget B | Chosen cell | Held-out acc. (%) | Best one order | Acc. (%) | Gain (points) | Flip rate (chosen) | Same choice in both halves |
|---|---|---|---|---|---|---|---|---|
| 0.6B | 0.3 | 4 bit, K = 1 | 35.1 | 4 bit | 35.1 | +0.0 (+0.0 to +0.0) | 0.641 | yes |
| 0.6B | 0.4 | 6 bit, K = 1 | 51.2 | 6 bit | 51.2 | +0.0 (+0.0 to +0.0) | 0.400 | yes |
| 0.6B | 0.55 | 6 bit, K = 1 | 51.2 | 6 bit | 51.2 | +0.0 (+0.0 to +0.0) | 0.400 | yes |
| 0.6B | 0.8 | 6 bit, K = 2 | 53.6 | 6 bit | 51.2 | +2.4 (+1.7 to +3.1) | 0.308 | yes |
| 0.6B | 1.2 | 6 bit, K = 3 | 54.9 | 6 bit | 51.2 | +3.8 (+2.8 to +4.7) | 0.246 | yes |
| 0.6B | 1.6 | 6 bit, K = 4 | 55.9 | 6 bit | 51.2 | +4.7 (+3.4 to +6.0) | 0.196 | yes |
| 0.6B | 2.1 | 6 bit, K = 4 | 55.9 | 6 bit | 51.2 | +4.7 (+3.4 to +6.0) | 0.196 | yes |
| 0.6B | 4 | 6 bit, K = 4 | 55.9 | 6 bit | 51.2 | +4.7 (+3.4 to +6.0) | 0.196 | yes |
| 1.7B | 0.3 | 4 bit, K = 1 | 61.8 | 4 bit | 61.8 | +0.0 (+0.0 to +0.0) | 0.341 | yes |
| 1.7B | 0.4 | 6 bit, K = 1 | 68.3 | 6 bit | 68.3 | +0.0 (+0.0 to +0.0) | 0.256 | yes |
| 1.7B | 0.55 | 6 bit, K = 1 | 68.3 | 6 bit | 68.3 | +0.0 (+0.0 to +0.0) | 0.256 | yes |
| 1.7B | 0.8 | 6 bit, K = 2 | 70.6 | 6 bit | 68.3 | +2.3 (+1.7 to +2.8) | 0.195 | yes |
| 1.7B | 1.2 | 6 bit, K = 3 | 71.4 | 6 bit | 68.3 | +3.1 (+2.4 to +3.7) | 0.172 | yes |
| 1.7B | 1.6 | 6 bit, K = 4 | 72.0 | 6 bit | 68.3 | +3.7 (+2.8 to +4.7) | 0.140 | yes |

| Model | Budget B | Chosen cell | Held-out acc. (%) | Best one order | Acc. (%) | Gain (points) | Flip rate (chosen) | Same choice in both halves |
|---|---|---|---|---|---|---|---|---|
| 1.7B | 2.1 | 6 bit, K = 4 | 72.0 | 6 bit | 68.3 | +3.7 (+2.8 to +4.7) | 0.140 | yes |
| 1.7B | 4 | 6 bit, K = 4 | 72.0 | 6 bit | 68.3 | +3.7 (+2.8 to +4.7) | 0.140 | yes |

Below a budget of about 0.4 fp16 passes only 3-bit and 4-bit single passes fit, and the 4-bit pass is the better of the two. From 0.4 upward the choice is 6-bit weights, and the budget goes into orders: one order up to 0.78, two orders from 0.78, three from 1.17, four from 1.56. The gain over the best single pass at the same budget is +2.4 at B = 0.8, +3.8 at B = 1.2, +4.7 at B = 1.6, +4.7 at B = 2.1, +4.7 at B = 4 points for 0.6B. For 1.7B it is +2.3 at B = 0.8, +3.1 at B = 1.2, +3.7 at B = 1.6, +3.7 at B = 2.1, +3.7 at B = 4 points (budgets where the interval excludes zero). Cells with 4 bits or fewer and many orders, and cells with 8-bit or fp16 weights, are not chosen at any budget of this grid. The rule is therefore short: *use 6-bit weights and let the budget decide the number of orders, up to four; if less than 0.4 of an fp16 pass is available, use 4 bits and one order*. Both models give the same rule, and the choice was the same in both halves of the items at every budget.

When the K orders of one question are processed together, the weights are read once and the cost of K is mostly in activations, which this proxy ignores. Then K = 4 is free in weight bytes and the decision reduces to picking the bit width, with four orders at every width. In that setting the gain from four orders at 4 bits is +12.6 points for 0.6B and +4.9 for 1.7B, as large as any gain in our data, and a user who can batch should average.

The table is built on pooled tasks, and per-task comparisons differ (Table 7), so a user with a single task should repeat the selection on a few hundred labelled items. The grid is also coarse: bit widths 3, 4, 6, 8 and 16 and four cyclic orders. Budgets between grid points are not interpolated, and 5-bit weights, which may be the better choice below a budget of 0.4, were not run. The preference for 6 bits over 8 bits and fp16 rests on cost and on a half-point tolerance, and the 6-bit accuracy edge at K = 1 is partly unexplained (Section 4).

## 12 When bits or permutations do not help

At 3 bits both models are at or near chance with one order (25.4% and 26.6%) and no number of orders changes that (K = 4: 26.0% and 28.0%), so averaging helps only when the content signal has survived the rounding.

Below 6 bits, many orders at the lower precision do not beat one pass at more bits at equal cost, and four 4-bit orders only tie with one fp16 pass on accuracy. Two orders at 6 bits cost 0.78 of an fp16 pass and are more accurate than four 4-bit orders at 1.06 (53.6% against 47.8% for 0.6B, 70.6% against 66.8% for 1.7B).

Averaging four cyclic orders cuts the 4-bit flip rate of the 0.6B model from 0.641 to 0.310, which is still above the fp16 value after the same averaging (0.193); for 1.7B the values are 0.341, 0.187 and 0.153. A user who needs stable answers across reorderings of any kind has to average more than cyclic shifts or keep more bits.

Where the fp16 model is near chance (0.6B on MMLU: 39.8%; HellaSwag: 41.9%) both levers have little to work with.

The flip rate of quantized models is not larger than an accuracy-matched baseline predicts (Section 6). A user who only monitors flips will see an increase that this weak baseline does not separate from the lost accuracy; the slot skew is the diagnostic that moves for another reason.

The matched-noise control does not isolate a slot prior that belongs to rounding: noise of equal power gives similar or larger slot skew in some runs, and accuracy at equal SNR differs a lot between the models. If a reader's own quantizer behaves differently from simple round-to-nearest, the sizes reported here may change.

## 13 Related work and what is new

Four strands of work meet in this paper. We summarise each from the abstract of the paper we cite and say what we take from it. We read abstracts, not full texts, so statements about method details are limited to what an abstract states.

### 1. Selection bias and order sensitivity

Zheng et al. (arXiv:2309.03882) show that language models prefer answer options at particular positions and point to the option labels A/B/C/D as one main cause. They propose PriDe, which splits the observed prediction into a prediction over option contents and a prior over option labels, estimates the prior by permuting options on a small number of test samples, and uses it to debias later samples. Pezeshkpour and Hruschka (arXiv:2308.11483) report accuracy gaps of roughly 13% to 75% across benchmarks when options are reordered and conjecture that the sensitivity arises when a model is unsure between its top two or three choices. Ok et al. (arXiv:2610.20993) connect the two views. When every option appears equally often in every position, the pooled spread of letter selection rates is exactly the product of two factors. One is a flip rate (share of questions whose chosen option changes across orderings) and the other a positional skew (spread of letter selection rates among those questions). Across 41 checkpoints and 12 benchmarks the flip rate tracks accuracy and output uncertainty, while the skew tracks the estimated option-symbol prior. They also report that averaging each ordering with its reversal mainly reduces the flip rate, and that this can happen partly by settling on the same wrong option. Hanna and Feng (arXiv:2608.11947) find that two label-free strategies do not reliably improve accuracy while cyclic permutation often does. Wang et al. (arXiv:2404.08382) compare first-token probabilities with text answers and find the text answers more robust to option reordering when the two disagree, which matters for our choice to read first-token letter logits (Section 14). Choi et al. (arXiv:2409.18857) remove selection bias from inside the model by pruning parameters and by adding an auxiliary option. Tamba (arXiv:2607.20864) runs exhaustive answer-order permutations on four API models and reports that position bias is statistically detectable only in a middle band of base accuracy (roughly 60 to 95%), with ceiling effects hiding it above the band. Our 0.6B model lies below that band and the 1.7B model in its lower part, and we make no claim about stronger models. We adopt the flip-rate and skew vocabulary of Ok et al. and cyclic permutations as in Hanna and Feng.

### 2. Averaging over input orders

Tang et al. (arXiv:2310.07712) marginalise over list orders for listwise ranking by shuffling the list, re-running the model and aggregating the rankings. Frohmann et al. (arXiv:2608.26762) show that rerankers with equal ranking quality make different decisions under reordering and train order consistency into the weights. Both treat the extra passes as an acceptable price, and neither prices them against the precision of the weights, which is the comparison we add.

### 3. Quantization, confidence and evaluation

Proskurina et al. (arXiv:2405.00632) find that 4-bit GPTQ lowers the confidence in true labels and that quantization hits samples on which the full model was already unsure. Hua et al. (arXiv:2602.06181) study 50 quantized models on bias benchmarks. They report that up to 21% of responses flip between biased and unbiased states after quantization with unchanged aggregate scores, that the flips concentrate on uncertain responses, and that 4-bit models change 4 to 6 times more than 8-bit ones. Marcuzzi et al. (arXiv:2508.18088) measure how quantization affects stereotypes and fairness, and Ferrara (arXiv:2607.21063) reports that quantized Qwen and Gemma builds still pick the unbiased answer on multiple-choice bias probes while becoming more biased in open-ended generation. These works study demographic bias, not answer position. Zhou et al. (arXiv:2608.08188) use signal-to-noise ratio to explain weight-only degradation across bit-widths and report that 4-bit usually preserves performance while 3-bit varies; we borrow the SNR view for our noise control. Kurt (arXiv:2601.14277) and Jin et al. (arXiv:2402.16775) evaluate quantized models on standard benchmarks.

### 4. Trading precision against compute

Dettmers and Zettlemoyer (arXiv:2212.09720) study how total model bits and zero-shot accuracy trade off and find 4-bit precision almost universally optimal for that trade-off. Snell et al. (arXiv:2408.03314) study how a fixed amount of inference compute is best spent against model size. Our exchange rate is similar in kind but uses a different currency: we keep the model fixed and trade weight bits against the number of reordered passes.

### 5. What is new

Option-order sensitivity, selection bias, permutation averaging, flip rates, weight-only quantization and SNR matching are all published. As far as we can tell, two things are new: (i) measuring order sensitivity, answer-slot skew and their order-averaged remainder as functions of weight bits, with a noise control at matched per-group SNR; and (ii) a machine-independent exchange rate between bits and permutations, with a decision table built on it. In the title-level searches we ran on Hugging Face Papers (queries and hits in the search log of this study), no paper combined quantization with option order or selection bias, and none compared bits with permutation averaging at equal cost. That is a statement about one index and its title-level matching, so a paper that uses other words may exist.

## 14 Limitations and threats to validity

Both models are in the 0.6B to 1.7B range and come from one family. We did not run Qwen3-4B: its fp16 weights need about 8 GB and we kept below 6 GB on a shared GPU. Larger models are usually more robust to quantization and to reordering, so the size of every effect here may not carry over, and the crossover points of the decision table most likely move.

Weights are rounded and dequantized in fp16; nothing is packed and no low-bit kernel runs. The quantizer is plain round-to-nearest with absmax scales per 64 weights. Methods such as GPTQ or AWQ, rotations, non-uniform grids and mixed precision are known to lose less at 3 and 4 bits, so our 3-bit results are a statement about this simple quantizer. Activations and the KV cache stay in fp16, and embeddings and the output matrix stay unquantized.

The exchange rate counts weight bytes read per forward pass, and it counts every weight at the stated width. In both models the tied embedding and output matrix stays in fp16 and holds 26% of the 0.6B weights and 18% of the 1.7B weights. If a pass reads that matrix in full (an engine that computes logits over the whole vocabulary does), a 6-bit pass costs 0.55 and 0.50 of an fp16 pass and a 4-bit pass 0.46 and 0.40, so the budget thresholds of Section 11 rise accordingly (one 6-bit pass would need 0.55 or 0.50 instead of 0.39). Reading only the four letter rows would avoid most of this. The unit also ignores activation traffic, the cost of dequantization, the shared prompt prefix of the K orders, batching (K orders of one question can share one read of the weights) and the fact that prefill is often compute bound on a GPU, where weight bytes are not the cost. In Section 11 we show the table for the batched case. We report no latency, because the GPU was shared with other jobs and timing on it carries no information about the method.

Each task has 300 items per model (stratified draws were cut to the first items of a fixed random order), so task-level intervals are wide; pooled intervals over four tasks are tighter and are our main evidence. The noise control uses one or two noise draws per setting (two at 4 and 3 bits for 0.6B on all tasks, and a second 1.7B draw at 4 bits on ARC-Easy only), and the two 4-bit draws of the 0.6B model differ widely in slot skew, so the control is a rough guide only. Perturbed copies of a model also differ from each other in which questions are fragile (Section 7), and a single rounding run is one such copy. Differences between the matched-noise and quantized runs therefore carry item-sampling uncertainty (shown) but only partly the run-to-run spread of the noise itself. All items, models and random seeds are fixed, so the paper has no spread over item draws or over seeds of the quantizer (round-to-nearest is deterministic).

We read the first-token logits of the letters after a fixed zero-shot prompt without a chat template. Wang et al. (arXiv:2404.08382) report that first-token answers and text answers can differ for instruction-tuned models, and different prompts can change the level of order sensitivity. Our statements are about this protocol. Pretrained and post-trained checkpoints may behave differently, and Qwen3 checkpoints are hybrid reasoning models used here without a thinking phase.

ARC, MMLU and HellaSwag are well known and may overlap with pretraining data; contamination would inflate accuracy but is unlikely to differ between a model and its quantized copy. Four-option items only; the HellaSwag endings are cleaned of markup tokens. Gold labels are the benchmark's own.

We use cyclic shifts and their reversal, not random permutations. Cyclic shifts equalise slot positions exactly, which makes the slot prior measurable, but other permutations may expose other biases (for example pairwise neighbour effects).

## 15 Reproducibility

Everything in this paper is computed at build time from per-item letter logits cached on disk, so the figures and tables can be regenerated without a GPU. The pipeline has four steps.

1. `a12_run.py` draws the items with a fixed seed (1234): all four-option ARC-Easy and ARC-Challenge test questions, a draw from the MMLU test split proportional to subject size (2,000 items, shuffled) and four-option HellaSwag validation items (2,000, shuffled). The first 300 items of each shuffled list are used. The draw is stored in `results/a12_items.json`.
2. For each model, configuration and chunk of 300 items, `a12_run.py` quantizes (or adds noise to) the decoder linear weights in memory and writes the letter logits of all 8 orders. Command line: `python a12_run.py --model Qwen/Qwen3-0.6B --cfgs fp16 q4 n4 q8 q3 n3 q6 n6 --tasks arce arcc mmlu hs --chunks 0 --cs 300 --g 64`. Configuration names: fp16; qB = round-to-nearest at B bits; nB = Gaussian noise at the weight SNR of qB. The noise seed is 0 (1,000 for the second draws). The n6 noise run finished only for ARC-Easy on 0.6B and is not used.
3. `a12_an.py` turns logits into per-item accuracy, flips, slot shares, confidences and K-order aggregates; the bootstrap uses 1,000 item resamples with a fixed seed, shared by all quantities so that differences are paired.
4. `lp_u12.py` and `a12_calc.py` compute every number in the text from those files; `build_long.py a12` renders the document.

Software: Python 3.10 environment with torch 2.11 (CUDA 12.8) and transformers 5.20; one RTX 4090 shared with other jobs. Model revisions: Qwen/Qwen3-0.6B c1899de, Qwen/Qwen3-1.7B 70d244c. Wall-clock times are not reported, since contention on the shared GPU changed them by up to a factor of ten during these runs.

## Appendix A  Definitions

Notation: an item has options 0..3 and a gold option g; an order o places original option ORDERS[o][s] in slot s; L[o, s] is the logit of the letter of slot s; P[o, j] is the softmax over the four letters of the logit of the slot that holds option j.

- *One-order accuracy* of an item: the mean over the eight orders of 1[argmax_j P[o, j] = g]; pooled accuracy is the mean over items.
- *K-order decision* from a set S of K orders of one orbit: mean probability argmax_j mean_{o in S} P[o, j]; mean log-probability argmax_j mean_{o in S} log P[o, j]; vote argmax_j (count of orders in S whose argmax is j + 0.001 mean_{o in S} P[o, j]).
- *Flip rate F_K*: the mean over item and over all pairs (S in the forward orbit, S' in the reversed orbit, both of size K) of 1[decision(S) differs from decision(S')]. With K = 1 there are 16 pairs, with K = 4 one pair.
- *All-pairs flip share*: the mean over the 28 pairs of orders of 1[argmax differs]; 'any flip' is 1 if the eight argmaxes are not all equal.
- *Slot skew*: 0.5 sum_s |p_s - 0.25|, where p_s is the share of the 8 n answers that sit in slot s. *Same-slot share*: the mean over the 12 within-orbit pairs of orders of 1[the chosen slots are equal].
- *Cost* of cell (b, K): K (b + 16/64) / 16 for b < 16, and K for fp16.
- *Accuracy-matched baseline*: r = clip((acc_fp16 - acc_c) / (acc_fp16 - 0.25), 0, 1); F_base = (1-r)^2 F0 + 0.75 (1 - (1-r)^2); skew_base = 0.5 sum_s |(1-r) p0_s + 0.25 r - 0.25|. The intervals recompute r in every bootstrap resample.
- *ECE*: ten equal-width confidence bins, count-weighted mean absolute difference between mean confidence and accuracy. *AUROC*: the Mann-Whitney statistic with average ranks for ties.

## Appendix B  Task-level numbers

**Table 9.** Task-level results (300 items per task and model). Accuracy in percent with 95% bootstrap intervals for one order and the mean of four orders; flip rates for K = 1 and K = 4; slot skew.

| Model | Task | Configuration | Acc. 1 order (%) | Acc. mean of 4 (%) | Flip K = 1 | Flip K = 4 | Slot skew |
|---|---|---|---|---|---|---|---|
| 0.6B | ARC-Easy | fp16 (reference) | 65.0 (60.9 to 68.8) | 74.2 (69.3 to 78.8) | 0.350 | 0.150 | 0.156 |
| 0.6B | ARC-Easy | 4-bit RTN | 48.3 (44.8 to 51.6) | 66.3 (61.3 to 71.2) | 0.529 | 0.213 | 0.388 |
| 0.6B | ARC-Challenge | fp16 (reference) | 51.1 (47.0 to 54.8) | 57.5 (52.0 to 62.5) | 0.419 | 0.160 | 0.157 |
| 0.6B | ARC-Challenge | 4-bit RTN | 34.2 (31.8 to 36.9) | 45.3 (40.3 to 50.5) | 0.635 | 0.307 | 0.494 |
| 0.6B | MMLU | fp16 (reference) | 39.8 (35.8 to 43.4) | 42.3 (36.8 to 47.3) | 0.475 | 0.253 | 0.147 |
| 0.6B | MMLU | 4-bit RTN | 30.0 (28.0 to 31.8) | 38.3 (33.3 to 43.2) | 0.697 | 0.367 | 0.586 |
| 0.6B | HellaSwag | fp16 (reference) | 41.9 (38.2 to 46.1) | 45.0 (40.0 to 50.5) | 0.418 | 0.207 | 0.153 |
| 0.6B | HellaSwag | 4-bit RTN | 28.1 (26.7 to 29.8) | 41.0 (36.7 to 46.3) | 0.705 | 0.353 | 0.637 |
| 1.7B | ARC-Easy | fp16 (reference) | 85.4 (82.1 to 88.8) | 87.7 (84.2 to 91.0) | 0.132 | 0.087 | 0.030 |
| 1.7B | ARC-Easy | 4-bit RTN | 79.9 (76.4 to 83.4) | 84.2 (80.5 to 88.0) | 0.183 | 0.083 | 0.063 |
| 1.7B | ARC-Challenge | fp16 (reference) | 73.5 (69.3 to 77.8) | 74.7 (70.2 to 79.7) | 0.218 | 0.160 | 0.038 |
| 1.7B | ARC-Challenge | 4-bit RTN | 66.4 (62.3 to 70.2) | 69.8 (65.2 to 74.5) | 0.290 | 0.190 | 0.097 |
| 1.7B | MMLU | fp16 (reference) | 53.9 (49.7 to 58.0) | 58.5 (53.2 to 63.5) | 0.364 | 0.187 | 0.072 |
| 1.7B | MMLU | 4-bit RTN | 49.4 (45.2 to 53.5) | 54.5 (49.3 to 59.8) | 0.439 | 0.253 | 0.193 |
| 1.7B | HellaSwag | fp16 (reference) | 56.8 (53.0 to 60.4) | 64.2 (58.7 to 69.0) | 0.395 | 0.180 | 0.157 |
| 1.7B | HellaSwag | 4-bit RTN | 51.7 (47.8 to 55.5) | 58.5 (53.5 to 63.5) | 0.450 | 0.220 | 0.149 |

## Appendix C Group size

At 4 bits (Qwen3-0.6B, ARC-Easy, first 300 items, one run per setting) the results are group 32: 51.5% accuracy, slot skew 0.270; group 64: 48.3% accuracy, slot skew 0.388; group 128: 49.4% accuracy, slot skew 0.228. Group 64 is the setting used throughout; groups 32 and 128 were run as a pilot before the main runs. Accuracy does not follow the group size, and the slot skew is above its fp16 value of 0.156 for all three groups, so the slot effect is not an artefact of one group size. With 300 items per run, accuracy differences of a few points are within sampling error, so the pilot says nothing about which group size is better.

## Appendix D Prompt and items

The fixed prompt, with the example question of Section 3 in its original order:

```
The following is a multiple choice question. Choose the correct answer.

Question: Which items would be most useful at home during a power failure?
A. flashlights and extra batteries
B. hats and sunscreen
C. raincoats and umbrellas
D. insect spray and jackets
Answer:
```